\documentclass[preprint,3p,times,twocolumn]{elsarticle}

\usepackage{amssymb}
\usepackage{lipsum}
\usepackage{amsthm}
\usepackage{xurl}

\usepackage{comment}
\usepackage{balance}
\usepackage{lineno}
\usepackage{multirow} 
\usepackage{float}
\usepackage{graphicx}
\usepackage{placeins}
\usepackage{pgfplots}
\pgfplotsset{compat=1.18}
\usepackage{tikz}
\usepackage{enumitem} 
\usepackage{amsmath}
\usepackage{algorithm}
\usepackage{algpseudocode}
\usepackage{cleveref}
\usepackage{booktabs}
\usepackage{listings}
\usepackage{outlines}
\usepackage{tikz}
\usetikzlibrary{shapes.geometric}
\usetikzlibrary{arrows.meta}
\usepackage{colortbl}
\usepackage[table]{xcolor}
\usepackage{soul}
\usepackage{colortbl}
\usepackage{graphicx}
\usepackage{booktabs}
\usepackage{siunitx}
\usepackage{subcaption}
\usepackage{tikz}
\usetikzlibrary{shapes,arrows}
\usetikzlibrary{shapes.geometric}
\usepackage{pgfplots}
\usepackage{pgfplotstable}
\usepackage{xcolor,colortbl}
\usepackage{pgf} 
\usepackage{amssymb}
\usepackage{pifont}
\usepackage{arydshln}
\usepackage{afterpage}
\usepackage[percent]{overpic}

\newcommand{\cmark}{\ding{51}}
\newcommand{\xmark}{\ding{55}}

\definecolor{inputcolor}{RGB}{255,0,0}
\definecolor{gtcolor}{RGB}{0,128,0}
\definecolor{predcolor}{RGB}{255,255,0}

\journal{Pervasive and Mobile Computing}

\begin{document}

\begin{frontmatter}





\title{Fusion of Pervasive RF Data with Spatial Images via Vision Transformers for Enhanced Mapping in Smart Cities}



\author[label1,label2]{Rafayel Mkrtchyan}
\ead{rafayel.mkrtchyan@ysu.am}
\author[label1,label2]{Armen Manukyan}
\ead{armen@yerevann.com}
\author[label1,label2]{Hrant Khachatrian} 
\ead{hrant.khachatrian@ysu.am}
\author[label3]{Theofanis P. Raptis\corref{cor1}}
\ead{theofanis.raptis@iit.cnr.it}

\affiliation[label1]{organization={Yerevan State University},
            addressline={Alex Manoogian 1}, 
            city={Yerevan},
            postcode={0025}, 
            country={Armenia}}
\affiliation[label2]{organization={YerevaNN},
            addressline={Alex Manoogian 1}, 
            city={Yerevan},
            postcode={0025}, 
            country={Armenia}}
\affiliation[label3]{organization={Institute of Informatics and Telematics, National Research Council},
            addressline={via G. Moruzzi 1}, 
            city={{Pisa}},
            postcode={56124}, 
            country={Italy}}

\cortext[cor1]{Corresponding author: T.~P.~Raptis}
\begin{abstract}
Accurate environment mapping is an important computing task for a wide range of smart city applications, including autonomous navigation, wireless network operations and extended reality environments. On the one hand, conventional smart city mapping techniques, such as satellite imagery, LiDAR scans, and manual annotations, often suffer from limitations related to cost, accessibility and accuracy. On the other hand, open-source mapping platforms, such as OpenStreetMap, have been widely utilized in artificial intelligence (AI) applications for environment mapping, serving as a source of ground truth. However, human errors and the evolving nature of real-world environments introduce biases that can negatively impact the performance of neural networks trained on such data. In this paper, we present a deep learning-based approach that integrates the DINOv2 architecture to improve building mapping by combining (possibly erroneous) maps from open-source platforms with pervasive radio frequency (RF) data collected from multiple wireless user equipments and base stations. Unlike prior methods, our approach leverages a vision transformer-based architecture to jointly process both RF and map modalities within a unified framework, effectively capturing spatial dependencies and structural priors for enhanced mapping accuracy. For the evaluation purposes, we employ a synthetic dataset co-produced by Huawei. To address the challenges associated with real-world data imperfections, we introduce controlled noise to its RF data so as to simulate real-world conditions. Additionally, we develop and train a model that leverages only aggregated path loss information to tackle the mapping problem. We measure the results according to three performance metrics: the Jaccard index (intersection over union, IoU), the Hausdorff distance, and the Chamfer distance. Our design achieves a macro IoU of 65.3\%, significantly surpassing (i) the erroneous maps baseline, which yields 40.1\%, (ii) an RF-only method from the literature, which yields 37.3\%, and (iii) a non-AI fusion baseline that we designed which yields 42.2\%. The comparative evaluation highlights the limitations of relying solely on RF data or on spatial data, as well as the effectiveness that AI can have on fusing data towards enhancing smart city mapping accuracy. We further validate our method on real-world data from the Oslo region, complementing the synthetic evaluation with a real deployment setting, where our best fusion model reaches 64.9\% macro IoU. We additionally outline a strategy for deploying the model over larger areas by tiling the region with overlapping windows.
\end{abstract}




\end{frontmatter}





\section{Introduction} \label{sec::intro}

\emph{Smart cities}, characterized by their \emph{pervasive integration} of digital technologies \cite{CESARIO2022101687} and interconnected systems \cite{BERALDI2020101221}, face unique challenges in accurately capturing and updating the physical and dynamic characteristics of urban spaces. \emph{Environment mapping} is the task of predicting the spatial layout of an area by identifying and \emph{localizing key structures} such as buildings, roads, and other physical obstacles. This problem is particularly challenging in urban and dynamic environments of smart cities, where structures may be missing, misaligned, or, in  general, \emph{inaccurately represented} in existing maps. Traditional mapping approaches rely on satellite imagery, LiDAR scans, or manual annotations, each of which comes with limitations in cost, accessibility, accuracy, or cost \cite{SOHN200743}. Accurate environment mapping is essential for numerous applications, including autonomous navigation, wireless network optimization, augmented reality, and virtual reality. 

A widely adopted resource in deep learning for environment mapping is a set of available \emph{open-source mapping platforms}, such as OpenStreetMap (OSM) \cite{OpenStreetMap}. These platforms have been extensively used as a source of ground truth for tasks such as building and road segmentation and change detection \cite{aerial1,chen2023land,laddha2016map}. However, several studies \cite{Brovelli2018,Herfort2023,Hecht2013} have highlighted \emph{inherent inaccuracies} and \emph{incompleteness} within open-source mapping platforms, including missing, misaligned, or outdated building structures. To mitigate these inaccuracies, various methods have been proposed for \emph{correcting such errors} using satellite imagery and deep learning techniques \cite{vargas2019correcting,li2022improving,li2023rethink}.

\emph{Radio-based information} has been extensively used as a low-cost alternative across a wide range of applications, mostly employing analytical methodologies, including \emph{wireless localization} of user devices \cite{locadhoc,locbds,loc1,loc2,loc3,loc4,mukherjee2019losi} and \emph{radio map prediction} \cite{indoorpathloss2,indoorpathloss4,indoorpathloss7,indoorpathloss8,outdoorpathloss2,outdoorpathloss3,pathlosstransformer,pathlossunet}
. In recent years however, \emph{deep learning} has emerged as a powerful tool for \emph{generalizing} across unseen environments in these domains. Various neural architectures have been explored, ranging from convolutional neural networks (CNNs) \cite{lecun1998gradient} to vision transformers (ViTs) \cite{vit}. Unfortunately, when it comes to RF signals, \emph{data scarcity} remains a significant challenge when training neural networks for radio-based tasks. To address this limitation, very recently, researchers have managed to produce \emph{synthetic datasets} via large-scale radio signal simulators \cite{wair-d,3gpp,sionna}, enabling scalable and controlled data generation. 

\paragraph{Novelty}
To the best of our knowledge, this is the first work to jointly fuse \emph{open-map priors} (OSM-style errors) with \emph{pervasive RF measurements} from multiple UE–BS pairs in a unified \emph{vision-transformer} framework \emph{for building-footprint refinement on WAIR-D under explicit OSM-style corruptions and realistic RF noise}. Unlike prior map-only correction or RF-only reconstruction, we turn RF propagation cues into a corrective signal for map inaccuracies (missing, shifted, or simplified buildings) while remaining robust to realistic RF noise and map corruption. We validate on the WAIR-D synthetic dataset \cite{wair-d} with 1,000 test environments, using empirically grounded corruptions and RF noise settings derived from the OSM error literature \cite{Brovelli2018}, \cite{Hecht2013} and \cite{ruble2018wireless} (28 GHz and 78 GHz).

\paragraph{Contributions} This paper provides the following contributions:

\begin{itemize}[leftmargin=*,nosep]
  \item \emph{Unified RF–map fusion via ViTs.} We introduce MapRadioFormer, a DINOv2-based encoder–decoder that jointly processes incomplete maps and structured RF tokens as a \emph{single} tokenized sequence, enabling cross-modal spatial learning. For this purpose, we define two RF granularities: \emph{R1} (path-level AoA/AoD/ToA per propagation path) and \emph{R2} (aggregated multi-band path loss at \{2.6, 6, 28, 60, 100\}\,GHz)(\Cref{sec::main-method}, \Cref{fig:dino++}).

  \item \emph{Realism for training and testing.} We design a map-corruption pipeline guided by OSM error statistics—removals (57\%), positional shifts (mean 1.46\,m; derived from {\Cref{tab:building-shift}}), and geometric simplification (25\%)—reflecting common open-map inaccuracies (\Cref{sec:corruption}). We also inject empirically grounded angular noise into R1 features at 28 GHz and 78 GHz following {\cite{ruble2018wireless}} (\Cref{sec:angle-noise}).

  \item \emph{Empirical gains across metrics.} On WAIR-D, our best fusion model (MapRadioFormer$_{\text{Map+R1}}$) achieves \textbf{65.3\% macro IoU}, surpassing (i) the corrupted-map baseline (40.1\%), (ii) an RF-only CLIP+UPerNet reproduction (37.3\%), and (iii) a non-learning fusion baseline (42.2\%). It also substantially lowers \emph{boundary} errors vs. the best RF-only model (R1): Hausdorff \textbf{34.0\,m} vs. 45.8\,m and Chamfer \textbf{2.0\,m} vs. 4.0\,m(\Cref{tab:model-comparison}). Notably, fusing maps with \emph{R2} (no angle hardware) still attains \textbf{55.7\%} IoU, whereas \emph{R1} with noise reaches \textbf{63.2\%}, clarifying accuracy–cost trade-offs.

\end{itemize}

\paragraph{Paper organization}
\Cref{sec::rworks} reviews RF-based mapping and map correction; \Cref{sec::problem} formalizes binary refinement;
\Cref{sec::main-method} presents MapRadioFormer; 
\Cref{sec::dataset} details WAIR-D, the Oslo real-life dataset, OSM-style corruptions, and RF noise;
\Cref{sec::baselines} defines baselines; 
\Cref{sec::res} reports metrics, visuals (\Cref{fig:results,fig:comparative_viz}), large-area deployment (\Cref{sec:deployment-large}), group-wise analysis, and ablations; \Cref{sec::conclusion,sec:future} conclude and provide the future research directions.

\section{Related Works} \label{sec::rworks}

RF signals have been exploited for device localization \cite{locadhoc,locbds,loc1,loc2,loc3,loc4,mukherjee2019losi}. A notable contribution is using RF signals at city scale for heuristic real-world localization: the LoSI system of \cite{mukherjee2019losi} demonstrates large-scale location inference from FM signal integration and estimation, showing that even simple broadcast RF modalities carry substantial geographic information.

Recent advances in environmental mapping have also leveraged radio frequency (RF) information, leading to significant breakthroughs in both indoor and outdoor scenarios. The literature demonstrates a variety of approaches, including both classical and deep learning-based methodologies. The emergence of deep learning has provided new opportunities to tackle mapping problems using data-driven models. However, training deep neural networks requires large-scale datasets, which are often costly and time-intensive to collect or generate.

In \cite{mobileOFDM}, a mapping algorithm is introduced that employs mobile user equipment with multiple OFDM subcarriers for enhanced spatial information extraction. The work in \cite{indoorsar} presents a multi-look fusion approach to predict the internal structure of buildings using synthetic aperture radar (SAR) mounted on a vehicle moving along a building's exterior. Similarly, \cite{SIRE} adopts a comparable methodology, utilizing an ultrawideband synchronous impulse reconstruction radar to achieve environmental reconstruction. Beyond radar-based approaches, commercial WiFi signals have also been used to reconstruct the interior layout of buildings: {\cite{zhang2021building}} performs tomographic reconstruction of indoor building layouts from RSSI measurements of off-the-shelf WiFi devices.

A different strategy is explored in \cite{emreflectors}, where multiple transmitters and reflectors integrated within a single user device are used to localize environmental reflectors. This is accomplished via a hybrid-criterion-based Bayesian inference framework. For indoor mapping, \cite{indoorrfmap} proposes two distinct approaches: $RFMap\text{-}1$, which involves a mobile setup with two transmitters and two receivers, and $RFMap\text{-}2$, which relies on a fixed setup comprising one transmitter and six receivers. Both methods utilize RF propagation software to generate synthetic datasets, facilitating the training of generative adversarial networks (GANs).

Outdoor mapping methodologies have also been extensively explored. In \cite{3dwifirssiuav}, the authors use two unmanned aerial vehicles (UAVs) as a transmitter-receiver pair, leveraging WiFi signal RSSI readings to reconstruct a 3D representation of the environment. Additionally, \cite{reconstructioniwcmc} presents a deep learning-based outdoor reconstruction approach that utilizes angle of departure (AoD), angle of arrival (AoA), and time of arrival (ToA) measurements derived from synthetic data generated via ray-tracing simulations. At a larger spatial scale, {\cite{peng20223d}} reconstructs 3D city maps from signal-to-noise ratio measurements of low-earth-orbit communication satellites, while {\cite{wang2025bit}} focuses specifically on estimating building heights from cellular mobile signals in order to build 3D city maps.

A closely related line of work is Radiocycle \cite{zheng2022radiocycle}, a deep dual-learning framework that simultaneously predicts the building map and the radio map of an environment from pathloss measurements. However, Radiocycle relies on pathloss measurements that cover roughly 20\% of the available area of the environment, which is an unrealistic measurement density for real-world urban deployments, where pathloss values are only available at the sparse spatial locations of actual UE-BS links rather than over a dense fraction of the scene.

Deep learning on satellite/aerial imagery can extract building footprints \cite{aerial1,li2022improving} and increase accuracy of existing maps, but faces constraints: high-resolution (sub-meter to $\sim$2\,m) data are costly, while freely available $10\,\mathrm{m}$ imagery suffers from pixel mixing at building scale. Persistent cloud cover, update frequency, and tree canopy create gaps. An extreme example when satellite imagery fails is right after major natural disasters like earthquakes, when dust, clouds and night time can make accurate mapping impossible. Mapping methods can also leverage street cameras and phone-based data collection, but these often require either extensive infrastructure or active user participation, and might raise privacy and access concerns. RF signals provide a complementary modality: wireless propagation interacts with buildings, offering structural cues when optical data are unavailable, outdated, or occluded. Our approach leverages existing cellular infrastructure to \emph{complement} imagery-based mapping, avoiding most privacy concerns.

Unlike other RF-based methods, which primarily rely on signal-based inference for mapping, our approach integrates open-source mapping platforms like OpenStreetMap with RF data to enhance building mapping. By leveraging publicly available maps, we introduce an additional structural prior that aids reconstruction and mitigates the limitations of purely RF-based methods. Furthermore, we propose a vision transformer-based model, MapRadioFormer, capable of processing both map and RF modalities in a joint framework. Our study highlights the benefits of multi-modal data fusion, demonstrating its effectiveness in enhancing the accuracy of environment mapping methodologies.

\section{Problem Formal Definition} \label{sec::problem}

\begin{figure}[t]
    \centering
    \includegraphics[width=1\linewidth, trim ={12mm 10mm, 0mm, 15mm}, clip]{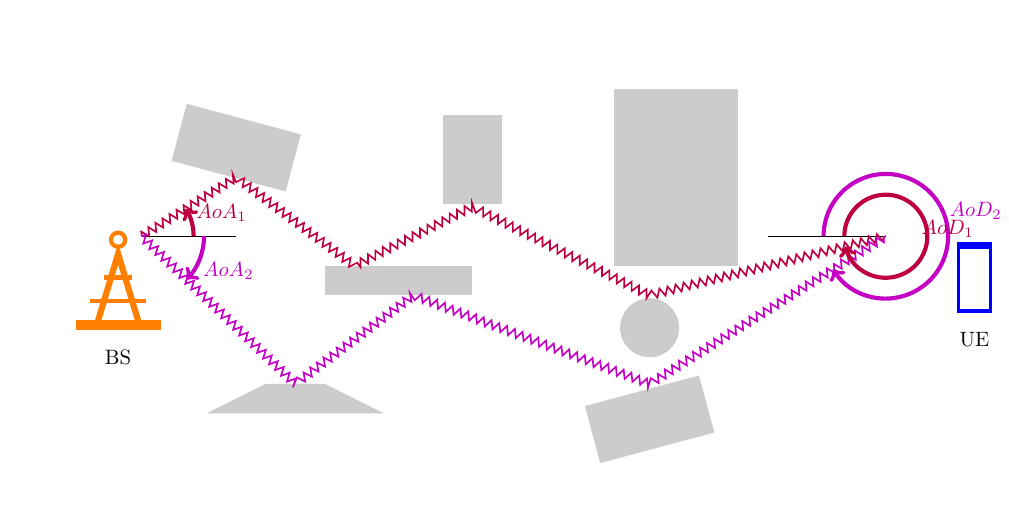}
    \caption{Visual representation of AoD and AoA. For a given user equipment (UE) to base station (BS) link, the angle of departure (AoD) is the angle at which the RF signal leaves the UE antenna, while the angle of arrival (AoA) is the angle at which the signal reaches the BS antenna.}
    \label{fig:aod-aoa}
\end{figure}

We consider an outdoor urban environment with multiple pervasive UEs and BSs. For each pair of UE and BS, multiple radio propagation paths may exist, each characterized by three key parameters: AoD, AoA and ToA. The AoD represents the direction at which the signal leaves the UE, while the AoA denotes the direction at which it reaches the BS. The ToA measures the duration required for the signal to travel from the UE to the BS along a specific propagation path. An example showcasing the AoD and AoA for two radio propagation paths can be seen in \Cref{fig:aod-aoa}. We consider two types of RF data.
\begin{itemize}
    \item R1: This representation captures detailed angular information, consisting of AoA, AoD and ToA. For each UE-BS pair, we select the five propagation paths with the shortest ToA, ensuring that the most significant signal paths are considered.
    \item R2: This representation is based on path loss measurements, which do not contain information about different radio propagation paths, and instead characterize the UE-BS pair with an aggregated number across multiple paths. Specifically, for each pair of UE and BS, we utilize path loss values corresponding to carrier frequencies of $\{2.6, 6, 28, 60, 100\}\mathrm{GHz}$.
\end{itemize}
Henceforth, we adopt the notation $\mathcal{R}$ to denote the RF data, which may correspond to either R1 or R2.

The building map of an outdoor environment is represented on a 2D grid of width $W$ and height $H$. Let $u \in \{0,1\}^{W \times H}$ be the true binary representation of an outdoor scene, or in other words a binary map, where $u(p)=1$ indicates that the pixel $p$ lies within a building footprint, and $u(p)=0$ represents a space without buildings. We also have an initial map $\tilde{u}\in \{0,1\}^{W\times H}$, which may be outdated, incomplete, or otherwise misaligned with the true environment. Our aim is to learn a function $f$
$$
    \hat{u} = f\bigl(\tilde{u},\mathcal{R}\bigr)
$$
that corrects the errors in $\tilde{u}$ by using relevant RF observations $\mathcal{R}$.

\begin{figure}[t]
    \centering
    \includegraphics[width=1\linewidth,trim={0 36mm 0 0mm},clip]{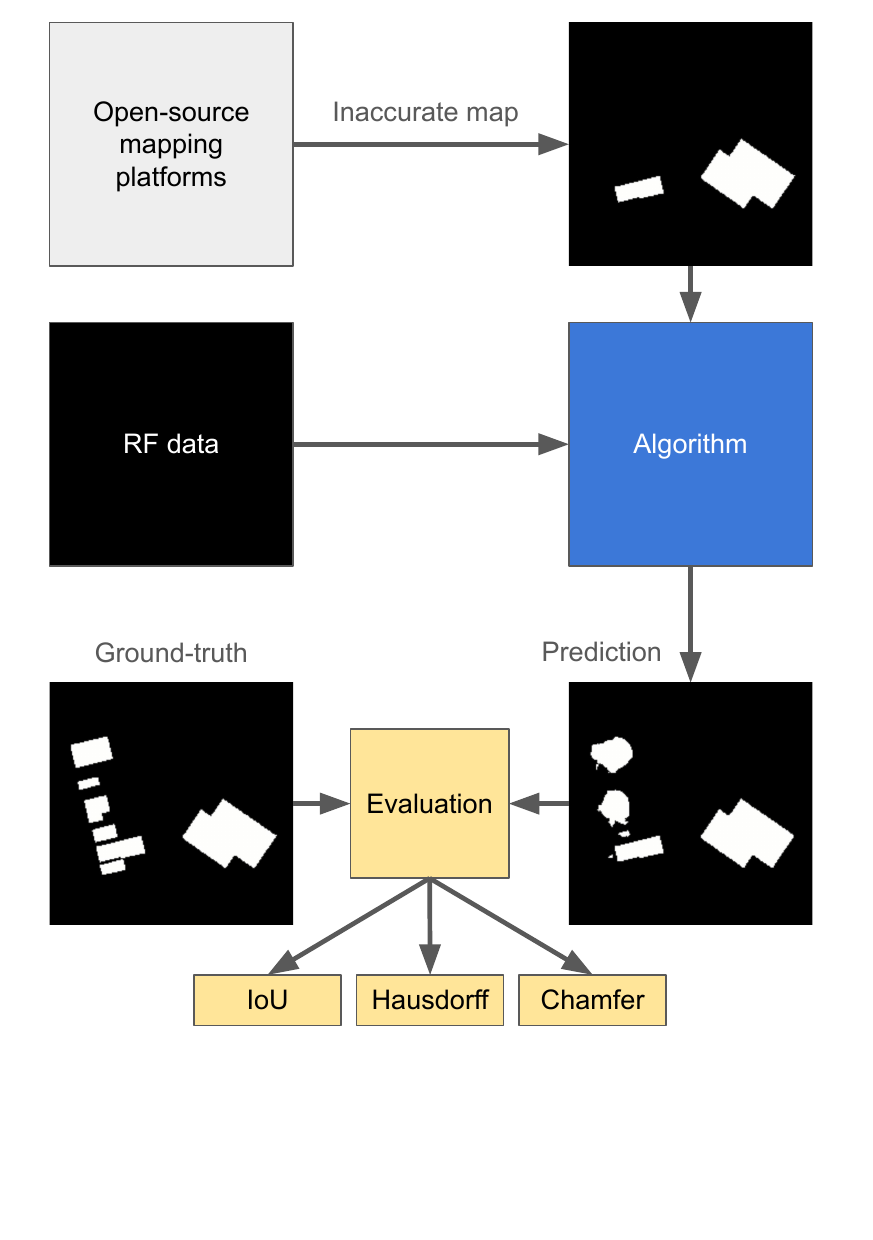}
    \caption{The overall structure of the study. Starting from the ground-truth binary building map, a synthetic corruption pipeline produces an inaccurate map that mimics typical open-source mapping errors. This corrupted map, together with RF observations between base stations and user equipments (angles, delays, or path-loss values), is provided as input to the algorithm, which predicts a refined binary building map. The refined map is then compared against the ground-truth map using IoU, Hausdorff and Chamfer metrics to quantify reconstruction quality.}
    \label{fig:flow}
\end{figure}

The overall methodology of this study is depicted in \Cref{fig:flow}. Initially, an inaccurate map is simulated to reflect the imperfections commonly found in open-source mapping tools. This map, along with RF data, serves as input to the algorithm, which aims to enhance building mapping. The output of the algorithm is subsequently compared against the ground-truth map, and its performance is quantitatively assessed using the IoU metric.

The goal is therefore to produce a \emph{refined binary map} $\hat u \in \{0,1\}^{W\times H}$ at test time; in our experiments we use $W=H=224$. In addition to binary accuracy (IoU), we report boundary-sensitive distances between $\hat u$ and $u$ in \Cref{sec::res}.

\section{Our Approach} \label{sec::main-method}

\subsection{RF Data Granularity}\label{sec:granularity}

In our setup, each pair of user equipment and base station $i, j$ provides radio observations reflecting signal propagation through the \emph{true} environment $$u$$. We consider two levels of measurement detail: \textbf{R1 Path-Level} and \textbf{R2 Aggregated Path-Loss}. Although both types originate from simulations in the WAIR-D dataset, their practicality differs in real-world scenarios.

\subsubsection{R1: Path-Level}
\label{sec:granularity-angles}

For each pair of user equipment $i$ and base station $j$ up to five strongest propagation paths $k$ are taken, each characterized by:
$$
   \bigl(\psi_{i,j,k},\;\phi_{i,j,k},\;\tau_{i,j,k}\bigr)
   \quad\text{for}\; k = 1,\dots,K_{i,j} \ (\le 5),
$$
where $\psi_{i,j,k}$ is the AoA, $\phi_{i,j,k}$ is the AoD, and $\tau_{i,j,k}$ is the ToA. We concatenate these path features into a 15-dimensional vector and then add the coordinates of the UE and BS, producing a 19-dimensional vector per pair. If less than 5 propagation paths are available we pad the corresponding vector with zeros. Although angle-based features can possibly boost our refinement performance, obtaining AoA/AoD data in practical deployments typically requires specialized antenna arrays or other direction-finding hardware, which can be complex and costly at large scales. 

\subsubsection{R2: Aggregated Path-Loss}
\label{sec:granularity-pathloss}

In contrast, we utilize path loss values from the WAIR-D dataset at five distinct frequency bands $\{2.6, 6, 28, 60, 100\}\mathrm{GHz}$. These five scalars are concatenated into a single vector, then augmented by the coordinates of the UE and BS, producing a  9-dimensional vector per pair. If path loss values are less than -160 (including $-\infty$), we replace them with -160. While R2 representation omits angle information, such path-loss measurements are comparatively easier to obtain in real deployments, requiring only signal power level readings rather than angle estimation. We train and evaluate our map-refinement approach using both granularities.

\subsubsection{$R2_{\mathrm{rich}}$: Real-World Measurements}
\label{sec:granularity-r2rich}

The real-world Oslo dataset described in \Cref{sec:oslo-dataset} exposes a richer set of per-pair radio measurements than the aggregated path loss used in R2. For each UE-BS pair we collect five scalar measurements: the received signal strength indicator (RSSI), the new-radio signal-to-interference-plus-noise ratio (NSINR), the new-radio reference signal received power (NRSRP), the new-radio reference signal received quality (NRSRQ), and the time of arrival (ToA). These five measurements only exist in the real-world Oslo dataset and are not available in the synthetic WAIR-D data. We refer to this granularity as $R2_{\mathrm{rich}}$.

\subsection{Encoding the map} \label{sec::enc-rf}

We represent the input $\tilde{u} \in \{0, 1\}^{224 \times 224}$ and the ground-truth map $u \in \{0, 1\}^{224 \times 224}$ maps as binary images of dimensions $224 \times 224$, where a value of 1 denotes the presence of a building, and a value of 0 indicates its absence.

\subsection{Encoding RF Signals}

RF observations for each UE–BS pair are initially represented as fixed-length raw feature vectors. For the detailed R1 configuration, this raw vector is 19-dimensional: constructed by concatenating up to five path-level features (AoA, AoD, and ToA) with the corresponding spatial coordinates. AoA and AoD initially in the $[-\pi, \pi]$ range are normalized to $[-1, 1]$. Following the methodology of \cite{wair-d}, ToA is normalized by dividing by the size of the environment in meters. 

In R2 configuration, path loss measurements from five frequency bands along the coordinates of UE and BS are concatenated into a 9-dimensional vector for each UE-BS pair. Path loss values are standardized with the mean $102$ and standard deviation $22$ for all carrier frequencies. The aforementioned numbers were empirically derived from the analysis on the training set. 

In the $R2_{\mathrm{rich}}$ configuration used on the real-world Oslo data, the five scalar measurements RSSI, NSINR, NRSRP, NRSRQ, and ToA defined in \Cref{sec:granularity-r2rich} are concatenated with the UE and BS coordinates into a 9-dimensional vector per UE-BS pair. Each of RSSI, NSINR, NRSRP, and NRSRQ is normalized using its mean and standard deviation, with statistics computed over the whole Oslo set: $-50.19$ and $9.91$ for RSSI, $-5.44$ and $9.92$ for NSINR, $-65.60$ and $12.06$ for NRSRP, and $-19.08$ and $6.45$ for NRSRQ. ToA is normalized by dividing by the crop side length in meters times $10$.

\subsection{Architecture}

\begin{algorithm*}[tb]
    \centering
    \begin{lstlisting}[
        language=Python,
        basicstyle=\ttfamily\small,
        numbers=left,
        numbersep=5pt,
        columns=fullflexible,
        keepspaces=true,
        showspaces=false,
        showstringspaces=false,
        showtabs=false,
        breaklines=true,
        keywordstyle=\color{blue},
        stringstyle=\color{red!80!black}\itshape,
        commentstyle=\color{gray},
        emph={return,int},
        emphstyle=\color{green!50!black}
    ]
def unpatch(
    tensor: torch.Tensor,
    h: int,
    w: int,
    channels: int,
    patch_size: int,
) -> torch.Tensor:
    """
    Args:
        tensor: tokens of shape (batch size, token size, h, w)
        h: image height / patch size
        w: image width / patch size
        channels: #channels in the image
        patch_size: patch size

    Returns:
        Unpatched image of shape
        (batch size, #channels, image height, image width)
    """
    return tensor.permute(
        0, 2, 3, 1
    ).reshape(
        -1, h, w, channels, patch_size, patch_size
    ).permute(
        0, 3, 1, 4, 2, 5
    ).reshape(
        -1, channels, patch_size * h, patch_size * w
    )\end{lstlisting}
    \caption{Python function to reconstruct an image from tokens. The input \texttt{tensor} has shape (batch size, token size, h, w), where the token size is the product of the number of image channels and the square of the patch size. Lines~20--21 permute the tensor so that the token dimension is moved to the end, yielding shape (batch size, h, w, token size). Lines~22--23 reshape the tensor, splitting the trailing token dimension into a channel dimension, a vertical patch dimension, and a horizontal patch dimension, giving shape (batch size, h, w, \#channels, patch size, patch size). Lines~24--25 permute the tensor again, moving the channel dimension to the front, the vertical patch dimension right after the grid height, and the horizontal patch dimension right after the grid width, producing shape (batch size, \#channels, h, patch size, w, patch size). Finally, lines~26--27 reshape the tensor by merging the vertical patch dimension with the grid height and the horizontal patch dimension with the grid width, resulting in the output image of shape (batch size, \#channels, image height, image width).}
    \label{alg:unpatch}
\end{algorithm*}

\begin{figure*}[t]
    \centering
    \includegraphics[width=1\linewidth,trim={0 0 5mm 0},clip]{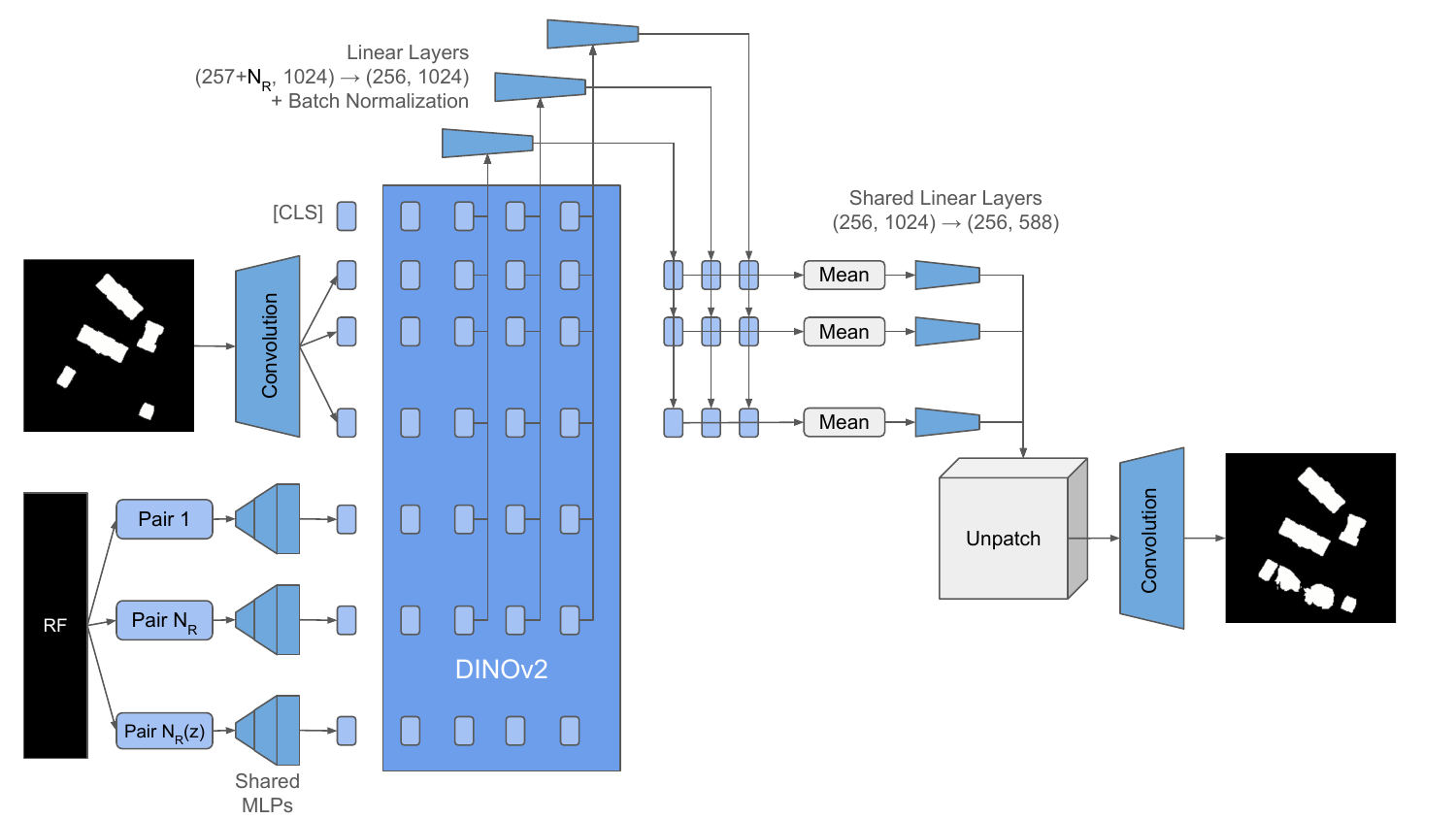}
    \caption{MapRadioFormer architecture. The model takes two inputs: a corrupted $224\times 224$ binary map and a set of per-pair radio information vectors. The corrupted map is lifted to a three-channel image by a $3\times 3$ convolution with same padding, while each radio vector is independently embedded by a shared two-layer MLP (256 and 1{,}024 ReLU units) into a 1{,}024-dimensional radio token; no positional embedding is added to radio tokens, so the network is invariant to UE-BS ordering. The map is split into $16\times 16=256$ patch tokens by the DINOv2 vision transformer (patch size $14\times 14$), and is processed jointly with the $N_R(z)$ radio tokens and a CLS token, giving $257+N_R(z)$ tokens of dimension 1{,}024 at every hidden layer. The first $257+N_R$ output tokens are passed through a linear projection to 256 tokens of size 1{,}024, followed by batch normalization, averaging across DINOv2 hidden layers, and a linear layer that maps each token to a vector of size $14\times 14\times 3=588$. These vectors are reshaped and tiled by the \emph{unpatching} operation in} \Cref{alg:unpatch} to form a $224\times 224\times 3$ feature image, which is finally passed through a $3\times 3$ same-padding convolution and a sigmoid to produce a per-pixel probability map; thresholding at $0.5$ yields the binary building prediction.
    \label{fig:dino++}
\end{figure*}

\begin{figure*}
    \centering
    \includegraphics[width=\linewidth]{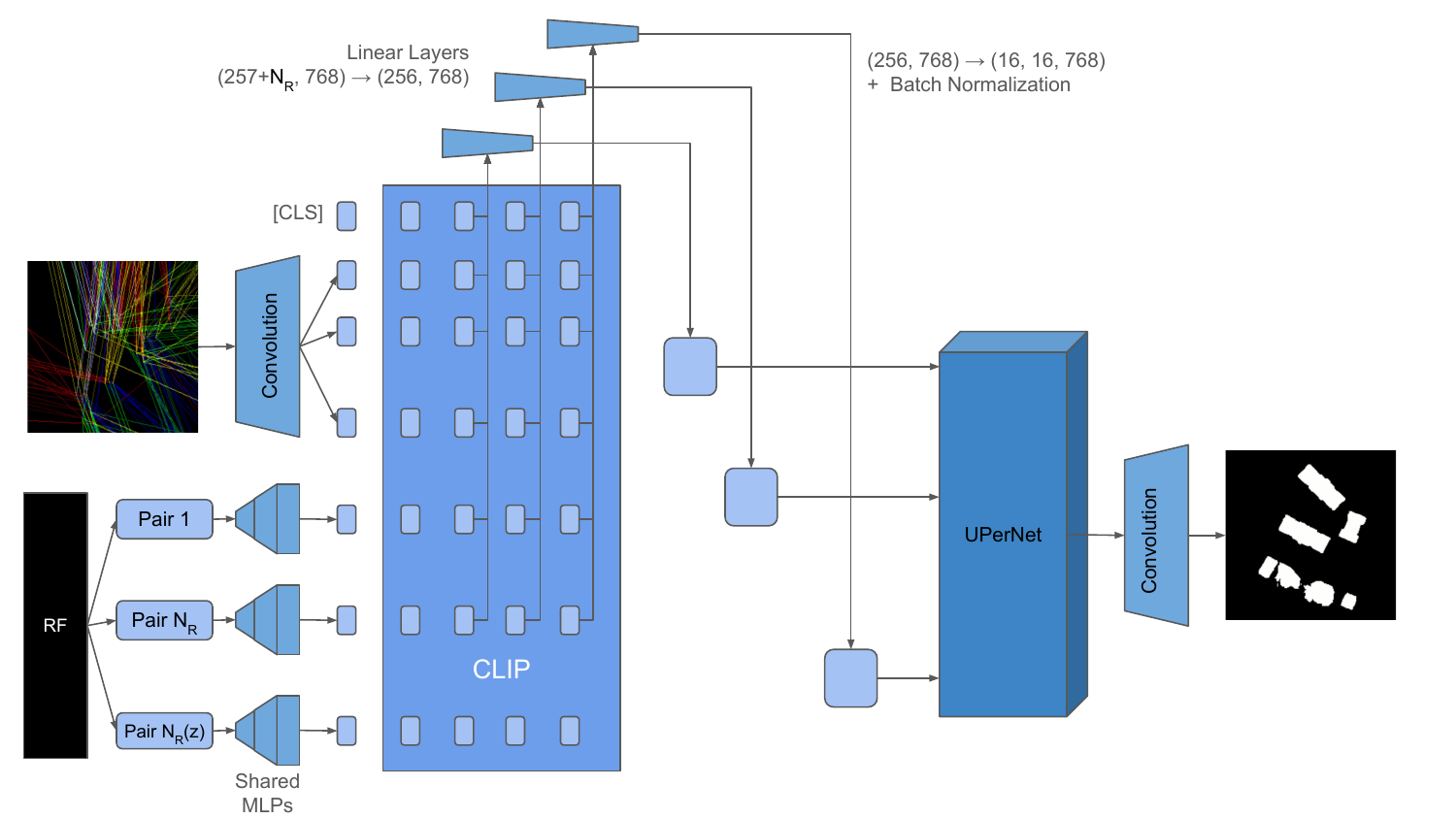}
    \caption{Architecture of the CLIP+UPerNet RF baseline \cite{reconstructioniwcmc}. The network jointly consumes RF information through radio ray maps and raw RF toknes, which are processed by a CLIP-based image encoder; the resulting features are then reshaped and passed to UPerNet decoder that produces the predicted building-footprint map. No map information is used as input.}
    \label{fig:clip-upernet}
\end{figure*}

Our proposed model processes a corrupted map, as described in \cref{sec:corruption}, alongside a set of radio information vectors, detailed in \cref{sec:granularity-angles,sec:granularity-pathloss,sec:angle-noise}. The corrupted map is first transformed into a three-channel image through a convolutional layer with a $3 \times 3$ kernel and same padding. Each radio information vector is independently processed using a shared multilayer perceptron (MLP) consisting of two hidden layers with 256 and 1,024 neurons, respectively, employing a ReLU activation function. The output size of this MLP is matched with the hidden vector dimension used in the DINOv2 vision transformer. We refer to the output of the MLP as \textit{radio tokens}. Given that the network must remain invariant to the ordering of UE-BS pairs, positional embeddings are not applied to the radio tokens.

Subsequently, both the three-channel image and the radio tokens are input to the DINOv2 vision transformer \cite{dinov2}. Let $N_R(z)$ denote the number of radio tokens for a given training sample $z$ and $\min N_R$ represent the minimum number of radio tokens required for the network to function. Given that the input map has a resolution of $224 \times 224$ and DINOv2 employs a patch size of $14 \times 14$, the image is decomposed into $\frac{224 \times 224}{14 \cdot 14} = 256$ tokens. When combined with the radio tokens and the CLS token, the model processes a total of $257 + N_R(z)$ tokens. Each hidden layer of DINOv2 generates $257 + N_R(z)$ tokens, each of size 1,024.

The first $257 + N_R$ tokens are passed through a linear transformation that reduces them to 256 tokens of size 1,024. This is followed by batch normalization to enhance training stability. The model then averages token representations across different hidden layers, resulting in 256 vectors of size 1,024. These vectors are subsequently transformed through an additional linear layer to produce 256 vectors of size $14 \times 14 \times 3 = 588$, allowing for an \textit{unpatching} operation. The vectors are then properly reshaped, aligning each token to its corresponding patch position, thereby reconstructing an image of dimensions $224 \times 224 \times 3$. This process is referred to as \textit{unpatching}. The Python code implementation for the unpatching operation utilizing the PyTorch library is presented in \Cref{alg:unpatch}. When calling the function, the parameters are set as follows: $\texttt{h}$=$\texttt{w}$=16, $\texttt{channels}$=3, and $\texttt{patch\_size}$=14.

To generate the final output, a convolutional layer with a $3 \times 3$ kernel and same padding is applied, producing a single-channel output image. A sigmoid activation function is then used to obtain a probability map, which is thresholded at 0.5 to derive a binary prediction. We refer to the resulting model as MapRadioFormer. The overall architecture of the network is illustrated in \Cref{fig:dino++}.

\subsection{Data augmentation}
\label{sec:aug}

To mitigate overfitting and enhance the diversity of training samples, we employ three straightforward data augmentation techniques:

\textbf{Flipping.} We apply both vertical and horizontal flipping to the input and output maps. To ensure consistency, the positions of base stations and user equipment are adjusted accordingly.
    
\textbf{UE-BS Subsampling.} For the synthetic WAIR-D data, for each training sample $z$, we uniformly sample a number $N_R(z)$ from the range $[N_R, 150]$. Subsequently, we randomly select $N_R(z)$ BS-UE pairs from the total set of 150 available pairs. For the real Oslo data, some samples contain many UE-BS pairs. Therefore, for each training sample $z$, which has $N_{pairs}$ available pairs, we sample $N_R(z)$ from the range $\Bigl[\min(300, \max(N_R, \frac{N_{\mathrm{pairs}}}{2})), \min(300, N_{\mathrm{pairs}})\Bigr]$, and subsequently select $N_R(z)$ random pairs. During inference, we used 300 random pairs.

\textbf{On-the-fly corruptions.} During training, the corruption process outlined in \cref{sec:corruption} is applied dynamically. Specifically, in each epoch, the map is corrupted using randomly selected parameters, ensuring variability in the corrupted inputs throughout the training process. 

\subsection{Severe corruptions during training}
\label{sec:severity}

To enhance training robustness, we investigate the impact of varying levels of map corruption severity during training. We define three levels of severity as follows:  

\begin{itemize}  
    \item \textbf{Level 1.} In this setting, the corruption parameters applied during training are identical to those used in the test set. Specifically, buildings are retained with a probability of $100\% - 57\% = 43\%$, the building shift is sampled from the distribution provided in \cref{tab:building-shift}, and simplification is applied to $25\%$ of the buildings.  

    \item \textbf{Level 1.5.} Here, the severity of corruption during training is increased by a factor of $1.5$ compared to the test set. Consequently, $1.5$ times fewer buildings are retained, leading to a building removal rate of $100\% - \frac{43\%}{1.5} \approx 100\% - 28.7\% = 71.3\%$. The building shift values are sampled from \cref{tab:building-shift} and scaled by a factor of $1.5$, while the simplification rate is increased to $1.5 \times 25\% = 37.5\%$.  

    \item \textbf{Level 2.} Similar to Level 1.5, this setting applies a corruption severity factor of $2$ during training. As a result, the building removal rate increases to $100\% - \frac{43\%}{2} = 100\% - 21.5\% = 78.5\%$, the building shift values from \cref{tab:building-shift} are doubled, and the simplification rate is raised to $2 \times 25\% = 50\%$.  
\end{itemize}

\subsection{Training objective}

For neural network optimization we adopt the Dice loss, $\mathcal{L}_{\mathrm{Dice}}(a,b) = 1 - \frac{2\,|a \wedge b|}{|a| + |b|}$, as a differentiable counterpart to IoU. Similar to IoU, Dice loss is bounded within $[0,1]$, where 0 represents a perfect match between maps, and 1 indicates no overlapping foreground pixels. Our objective is to minimize the expected Dice loss over all training samples ${(u, \tilde{u}, \mathcal{R})}$, where $\mathcal{R}$ represents either R1 or R2 measurements. Formally, the training objective is to find a function $f$ that minimizes the following objective function:
$$
  \min_f \mathbb{E}\Bigl[\mathcal{L}_{\mathrm{Dice}}\bigl(u, f(\tilde{u},\mathcal{R})\bigr)\Bigr].
$$

\subsection{Evaluation metrics}
\label{sec:eval-metrics}

For two binary maps $a$ and $b$, we use intersection-over-union, $\mathrm{IoU}(a,b) = \frac{|a \wedge b|}{|a \vee b|}$ as our main objective. IoU takes values in the range $[0,1]$ indicating how close the predicted and actual binary maps are, 0 being totally dissimilar and 1 being exactly the same. Having the evaluation metric in the $[0,1]$ range gives us interpretability in terms of performance. We aim to maximize the expected IoU over unseen examples $\{(u, \tilde{u}, \mathcal{R})\}$. Formally, our goal is to find a function $f$, that optimizes the following objective
$$
  \max_f \mathbb{E}\Bigl[IoU\bigl(u, f(\tilde{u},\mathcal{R})\bigr)\Bigr].
$$
Here, the expectation $\mathbb{E}[\cdot]$ is taken over the distribution of unseen samples, ensuring that the function $f$ generalizes well to unseen data. The function $f$ takes as input the corrupted map $\tilde{u}$ and the corresponding radio measurements $\mathcal{R}$ and produces a predicted map. The objective encourages $f$ to generate predictions that maximize the IoU with the ground truth $u$, thereby improving the overall mapping accuracy.

Since IoU is invariant to the size of the environment, we additionally report Hausdorff \cite{hausdorff_grundzuge_1914} and Chamfer \cite{10.5555/1622943.1622971} distances, both of which are directed metrics for evaluating spatial discrepancies between predicted and ground-truth maps. Let $P_u$ denote the set of points defining the polygons in the binary map $u$. Given two binary maps, $u_1$ and $u_2$, we define $d(p, P_{u_2})$ as the distance from a point $p \in P_{u_1}$ to the closest point in $P_{u_2}$. The Hausdorff distance is a directed metric that quantifies the maximum of such distances, measuring the worst-case deviation from one map to another. In contrast, the Chamfer distance computes the mean of these distances, thus, being less sensitive to outliers. These metrics are formally defined as follows:
\begin{equation*}
\begin{aligned}
    \overrightarrow{Hausdorff}(u_1, u_2) = 
    \max\limits_{p \in P_{u_1}}{d(p, P_{u_2})}
\end{aligned}
\end{equation*}
\begin{equation*}
\begin{aligned}
    \overrightarrow{Chamfer}(u_1, u_2) = 
    \frac{1}{|P_{u_1}|}\sum\limits_{p \in P_{u_1}}{d(p, P_{u_2})}
\end{aligned}
\end{equation*}

To obtain a symmetric evaluation, we aggregate the two directed metrics. Specifically, the overall Hausdorff distance is defined as the maximum of the two directed Hausdorff distances, while the Chamfer distance is computed as the average of its two directed counterparts:

\begin{equation*}
\begin{aligned}
    & Hausdorff(u_1, u_2) = \\
    & \max\bigl[
    \overrightarrow{Hausdorff}(u_1, u_2) \; , \; 
    \overrightarrow{Hausdorff}(u_2, u_1)
    \bigr]
\end{aligned}
\end{equation*}
\begin{equation*}
\begin{aligned}
    & Chamfer(u_1, u_2) = \\
    & \frac{1}{2} \times \bigl[
    \overrightarrow{Chamfer}(u_1, u_2) \; + \;
    \overrightarrow{Chamfer}(u_2, u_1)
    \bigr]
\end{aligned}
\end{equation*}

\subsection{Training details}

All models were trained on an NVIDIA DGX A100 system, utilizing two 40GB A100 GPUs with distributed data parallelism. For models trained on the synthetic WAIR-D data, training was conducted for 200 epochs. For models trained from scratch or fine-tuned on the real Oslo data described in \cref{sec:oslo-dataset}, optimization was performed over 400K training samples, with validation after every 10K samples. For the synthetic WAIR-D data, a learning rate scheduler was employed, which linearly increases the learning rate from $0$ to $2 \times 10^{-4}$ over the first 20 epochs, followed by a cosine decay to $0$. For the real Oslo data, we used a learning rate scheduler with linear warmup from $0$ to a peak learning rate of $10^{-4}$ over the first half of training, followed by a cosine decay to $0$ over the second half. For the synthetic WAIR-D data, the batch size was set to 15, as this was the maximum capacity that fit within the available GPU memory. For the real Oslo data, the batch size was set to 8. For all the experiments, the pretrained weights of DINOv2-L/14 ViT were used to initialize the DINOv2 module of the MapRadioFormer network. Following the training, the model checkpoint with the lowest validation loss was selected. To ensure reproducibility, the \texttt{seed\_everything} function from the PyTorch Lightning library was used with a fixed seed value of 0 for all the experiments.

\section{Dataset}
\label{sec::dataset}
\begin{table*}[t!]
    \centering
    \begin{tabular}{ccccccccc}
        \toprule
        \multirow{2}{*}{\textbf{Mean (m)}} & \multicolumn{8}{c}{\textbf{Percentile (m)}} \\
        \cmidrule{2-9}
        & 20\% & 40\% & 60\% & 80\% & 85\% & 90\% & 95\% & 100\% \\
        \midrule
        1.460 & 0.020 & 0.863 & 1.555 & 2.500 & 2.867 & 3.377 & 4.197 & 6.000 \\
        \bottomrule
    \end{tabular}
    \caption{Building shift distribution in meters. The values summarize the empirical building-location error distribution derived from \cite{Brovelli2018}, which quantifies positional mismatches between OSM building footprints and an authoritative reference dataset. The reported mean error is $1.46$~m, and the remaining columns list the 20\%, 40\%, 60\%, 80\%, 85\%, 90\%, 95\% and 100\% percentiles of the shift magnitude. We use this distribution in the map-corruption pipeline (see \cref{sec:corruption}): for every building, the magnitude of the positional shift is drawn from the CDF defined by linear interpolation between these percentiles, and the direction is sampled uniformly at random from $[0, 2\pi]$.}
    \label{tab:building-shift}
\end{table*}

In this section, we present the synthetic WAIR-D and real Oslo datasets utilized in our study. First, we provide a detailed description of the synthetic WAIR-D dataset in \cref{sec:wair-d}, which was generated through ray-tracing simulations and thus offers ideal angle information, including AoA, AoD and ToA. We then introduce the real Oslo dataset in \cref{sec:oslo-dataset}, which allows us to evaluate the proposed approach beyond the synthetic setting. However, acquiring such angle information in real-world scenarios requires specialized antennas, which may be costly.

Even with specialized antennas, perfect path-level information remains unrealistic, as real-world conditions introduce noise into the angle measurements, which varies depending on the carrier frequency. To reflect these real-world imperfections, we introduce noise into the angle data in \cref{sec:angle-noise}, following the methodology outlined in \cite{ruble2018wireless}. Indicatively, we do so for the angle data corresponding to carrier frequencies of 28 GHz and 78 GHz. Furthermore, open-source mapping tools such as OSM are known to exhibit inherent inaccuracies due to environmental changes and human-induced errors \cite{Brovelli2018, Hecht2013, Herfort2023}. These inaccuracies manifest as missing or misaligned buildings and insufficient detail in building representations. To replicate these mapping inaccuracies, we present an approach detailed in \cref{sec:corruption}.

\subsection{The WAIR-D dataset}
\label{sec:wair-d}
The WAIR-D dataset \cite{wair-d} contains 10,000 urban environments with scattered buildings based on maps of real cities, the locations of multiple base stations, and source nodes, along with radio link parameters. The environments are randomly sampled from more than 40 of the largest cities worldwide using the OSM service. However, information on real-life locations is not shared. Each environment is associated with a region-size value, since the spatial extent of the sampled regions varies across environments and is provided as part of the environmental metadata. The dataset comes with five channel frequencies $C \in \{2.6, 6, 28, 60, 100\}$ GHz. We followed the ``Scenario 1'' subset of the data where each urban environment contains $5$ base stations and $30$ user equipments. This produces $150$ BS-UE pairs per environment. BSs are placed at a fixed height of 6\,m and UEs at 1.5\,m, and both are equipped with omni-directional antennas. Radio propagation paths are generated by the 3D ray-tracing simulator PyLayers and include direct propagation, reflection, and diffraction, producing both line-of-sight and non-line-of-sight paths for each link. We used $L = 8,403$ maps as the training set, 596 as the validation set, and 1,000 as the test set. The data for the environment \#09117 was corrupt and, hence, was left out. WAIR-D also defines a second scenario (Scenario 2) with 100 environments, each containing a single BS and 10,000 densely dropped UEs. This scenario is used in our addtinal experiments on mobility in \cref{sec:mobile}. Note that all Scenario 2 environments are excluded from the training set.

\subsection{The Oslo dataset}
\label{sec:oslo-dataset}

\begin{figure*}[!t]
    \centering
    \includegraphics[width=0.85\textwidth]{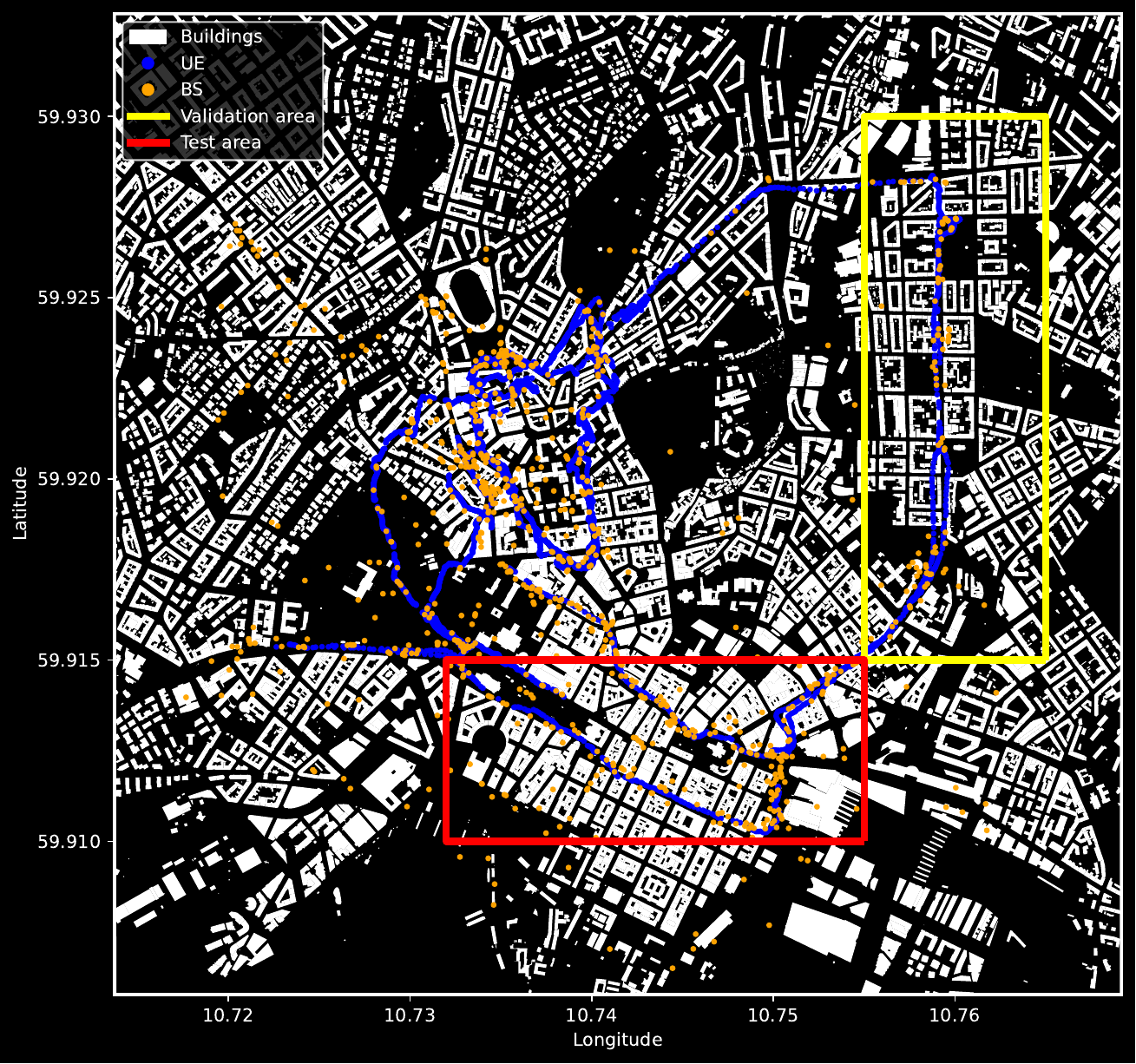}
    \caption{Geographic partition of the Oslo study area used to construct the real-life evaluation dataset. The outer rectangle corresponds to the full Oslo region from which crops are sampled, while the inner highlighted regions indicate the validation and test areas.}
    \label{fig:oslo-split}
\end{figure*}

To complement the synthetic WAIR-D data, we additionally evaluate our approach on the Oslo subset of the large-scale outdoor propagation dataset reported in~{\cite{s23094266}}. This real-life dataset supports only pathloss-based $R2$-type inputs, because each UE-BS pair is associated with a single aggregated pathloss value and no path-level AoA or AoD information is available. Moreover, the carrier frequencies associated with these Oslo measurements are not known to us.

For the spatial sampling procedure, we consider the Oslo region delimited by the following boundaries: north 59.94, south 59.90, east 10.77, and west 10.71. Within this region, we generate 1,000,000 square crops by sampling crop centers uniformly at random and sampling the crop side length uniformly from 300\,m to 500\,m. We make sure, that those crops contain at least 10 UE-BS pairs.

The dataset is then partitioned geographically. A crop is assigned to the test split only if it lies entirely within the test region delimited by north 59.915, south 59.910, east 10.755, and west 10.732. A crop is assigned to the validation split only if it lies entirely within the validation region delimited by north 59.930, south 59.915, east 10.765, and west 10.755. A crop is assigned to the training split only if it lies entirely outside both the validation and test regions. Crops that do not satisfy any of these criteria are discarded. This procedure yields 426,914 training crops, 59,238 validation crops, 41,230 test crops, and 472,618 discarded crops. From these retained subsets, we use the first 1,000 test crops as the final test set and the first 1,000 validation crops as the final validation set. The resulting geographic partition is illustrated in Figure~\ref{fig:oslo-split}.

Since the Oslo dataset provides only a single pathloss value for each UE-BS pair, we replicate this value multiple times so that the pathloss component matches the dimensionality of the WAIR-D $R2$ feature vector before appending the UE and BS coordinates. To ensure compatibility with our WAIR-D $R2$ processing pipeline, for each Oslo UE-BS pair we construct a feature vector that follows the same structure as the WAIR-D $R2$ representation and normalize it using the same statistics. The corresponding building footprints are retrieved from OpenStreetMap. The map corruptions follow the same procedure used for the WAIR-D map corruptions and are generated on-the-fly for every crop in order to increase training diversity.

\subsection{Simulating real-life angle data}
\label{sec:angle-noise}

In the simulation, angle information is assumed to be perfectly known and independent of the carrier frequency, which does not hold in real-world scenarios. To more accurately reflect real-life conditions, Gaussian noise is introduced to both the AoA and AoD for each UE-BS pair. Following the methodology outlined in \cite{ruble2018wireless}, noise is added to the AoA and AoD to simulate realistic measurements for two carrier frequencies: 28 GHz and 73 GHz. We restrict the angular noise injection to these two carrier frequencies because they are the only ones for which the reference measurement study \cite{ruble2018wireless} reports empirically calibrated AoA and AoD noise characteristics that can be directly reused in our simulation. Specifically, for a 28 GHz carrier frequency, the noise added to the AoD is sampled from $\mathcal{N}(0, 8.5^\circ)$, while for the AoA, it is sampled from $\mathcal{N}(0, 10.5^\circ)$. For a 73 GHz carrier frequency, the noise added to the AoD follows $\mathcal{N}(0, 5.5^\circ)$, whereas the AoA noise is sampled from $\mathcal{N}(0, 8.5^\circ)$.

\subsection{Simulating map corruption} 
\label{sec:corruption}

Open-source mapping platforms inherently contain systematic errors when compared against ground-truth building footprints~\cite{Brovelli2018,Hecht2013,Herfort2023}. For example in OSM, three issues predominate: (i)~inaccurate building positions, (ii)~omitted buildings, and (iii)~lower detail of the edges of polygon shapes. Although OSM can also contain footprints that do not exist in reality (around $9\%$ false positives \cite{Hecht2013}), we focus on the more prevalent issues of missing and misaligned buildings, and do not introduce extra false footprints in our corruption approach, as this error type is relatively less common. 

We initially simulate the identified errors and incorporate them as inputs into our models. The methodologies employed to simulate each type of error are detailed below.  

\textbf{Positional shifts.} For each building on a given map, we sample a shift distance $$s$$ from an empirically derived distribution introduced in \cite{Brovelli2018}. We do a linear interpolation on the cumulative distribution function (CDF) to fill in the values between the percentiles mentioned in \Cref{tab:building-shift}. The direction of the shift is then selected uniformly at random from $$[0,2\pi]$$. The amount of shift is sampled individually for each building during training. However, for the test set all buildings within the same environment are shifted with the same amount, but in different directions. This allows us to group test samples according to their amount of shift for detailed evaluation \cref{sec::grev}.

\textbf{Missing buildings.}
Each building is removed with probability of $57\%$, reflecting OSM under-reporting rates observed in practice \cite{Brovelli2018}. 

\textbf{Polygon simplification.}
Each remaining building is approximated by its convex hull to simulate the geometric coarseness observed in OSM \cite{Hecht2013}. During the training phase, this simplification is applied individually to $25\%$ of the buildings. To enable the grouping of environments in \cref{sec::grev}, $25\%$ of the environments undergo simplification, where all buildings within the same environment are simplified, while the buildings in the remaining environments are kept as is.

\textbf{Fixing the test set.}
For evaluation, we select $1,000$ WAIR-D environments and corrupt each environment with the aforementioned procedure. We record the shift distance $s$, the removal fraction $r$, and the simplification status for each environment. 

\begin{figure}
    \centering
    \includegraphics[width=\linewidth]{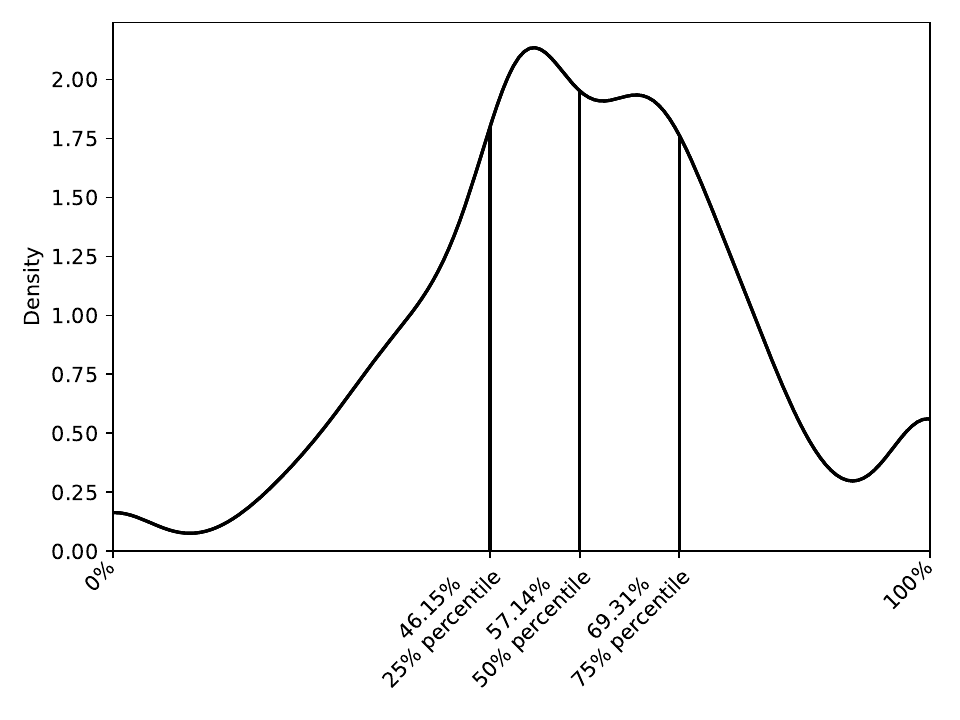}
    \caption{The distribution of the actual percentage of the buildings removed for the test set. Since each building is independently kept with probability $43\%$ (and removed with probability $57\%$), the effective per-environment removal rate fluctuates around its nominal value. This figure reports the distribution of the realized fraction of buildings that were actually removed by the corruption process, and is later used to bin the test set for the detailed evaluation reported in} \Cref{tab:detailed_eval}.
    \label{fig:removal-dist}
\end{figure}

In \cref{sec::grev}, we provide a detailed evaluation by grouping these $1,000$ test maps into $3\times 4\times 2$ bins, where the bins correspond to the groups:
\begin{outline}
    \1 \textbf{Shift range $s$:} 
      \2 $s < 1.46\,m$,\;
      \2 $1.46m \!\le\! s < 4.20\,m$,\;
      \2 $s \ge 4.20m$,
    \1[] where $1.46m$ and $4.20m$ shifts correspond to the mean and $95\%$ percentile.
    \1 \textbf{Removal rate $r$:} 
      \2 $r \leq 46.15\%$,
      \2 $46.15\% < r \leq 57.14\%$,
      \2 $57.14\% < r \leq 69.31\%$,
      \2 $r > 69.31\%$,
     \1[] where $46.15\%, 57.14\%$ and $69.31\%$ are $25\%, 50\%$ and $75\%$ percentiles of the distribution of the actual percentage of the buildings removed, the probability density function (PDF) of the distribution can be seen in \Cref{fig:removal-dist}.
    \1 \textbf{Simplification:} 
      \2 ``No'' (simplification is not applied),
      \2 ``Yes'' (simplification is applied).
\end{outline}

\section{Baselines} \label{sec::baselines}

\begin{algorithm*}[tb]
\caption{Non-AI Map Refinement}
\label{alg:refinement}
\begin{algorithmic}[1]
\Require 
   Initial map $\tilde{u} \in \{0,1\}^{224 \times 224}$;
   UE--BS sequences $\mathcal{S}$ (with LOS/NLOS labels and pair coordinates);
   max iterations $T$.
\Ensure 
   Refined map $\hat{u}$.

\State $\hat{u} \gets \tilde{u}$
\State $(V_{\text{nlos}},\,V_{\text{los}},\,\mathbf{p}) \gets \Call{AnalyzeViolations}{\hat{u},\,\mathcal{S}}$ \Comment{\textbf{p} contains the pixel coordinates of violation locations}
\For{$t \gets 1$ to $T$}
  \If{$|V_{\text{nlos}}|=0$}
     \State \textbf{break} 
  \EndIf

  \State $\mathbf{C} \gets \Call{RandomSample}{\mathbf{p},\,k=\min(|\mathbf{p}|, 40)}$
  \State $\mathcal{I} \gets \emptyset$  \Comment{candidate improvements list}

  \ForAll{$c \in \mathbf{C}$}
    \State $B \gets \Call{CreateRandomBuilding}{c}$ 
    \State $u' \gets \max(\hat{u},\,B)$ \Comment{pixelwise merge the new building}
    \State $(V'_{\text{nlos}},\,V'_{\text{los}},\,{-}) \gets \Call{AnalyzeViolations}{u',\,\mathcal{S}}$
    \If{$(|V_{\text{nlos}}'| + |V_{\text{los}}'| < |V_{\text{nlos}}| + |V_{\text{los}}|) \textbf{ and } (|V_{\text{los}}'| - |V_{\text{los}}| < 2)$}
       \State $\mathcal{I} \gets \mathcal{I} \cup \bigl\{(u',\,|V'_{\text{nlos}}|,\,|V'_{\text{los}}|)\bigr\}$
    \EndIf
  \EndFor

  \If{$\mathcal{I}=\emptyset$}
     \State \textbf{break} \Comment{no further improvement possible}
    \Else
   \State Sort $\mathcal{I}$ by $(|V_{\text{nlos}}| + |V_{\text{los}}|,\, |V_{\text{los}}|)$
\State $(\hat{u},\,|V_{\text{nlos}}|,\,|V_{\text{los}}|) \gets \mathcal{I}_1$
  \EndIf
\EndFor
\State \textbf{return} $\hat{u}$
\end{algorithmic}
\end{algorithm*}

Establishing robust baselines is essential to quantify the improvements introduced by our multi-modal fusion approach. We evaluate single-modality baselines to isolate the individual contributions of corrupted maps and RF signals. In addition, we introduce a non-AI fusion baseline as a lower bound, given the absence of state-of-the-art models that integrate both map and radio data for our task at hand. As shown in \cref{sec::res}, our best model leveraging both modalities significantly outperforms these baselines.

\subsection{Single-modality Baselines}
To evaluate the individual contributions of corrupted map inputs and RF signals, we consider two single-modality baselines: One that relies exclusively on the initial corrupted maps and another that utilizes only the available radio information in a given environment.

\paragraph{Corrupted Map Baseline}
This baseline measures the performance achievable by directly using the corrupted map $\tilde{u}$ without any refinement. By comparing $\tilde{u}$ with the ground-truth map $u$ over a test set of 1,000 WAIR-D environments, we obtain an average IoU of $40.1\%$. As detailed in \cref{sec:corruption}, the initial maps are affected by systematic errors quantified by positional shifts, building removal percentages, and polygon simplification. For that reason, we present the detailed evaluation based on the error groups of the corruptions in \cref{sec::grev}. This reference performance establishes a foundation for assessing improvements through additional modalities.
Moreover, we train a network exclusively on $\tilde{u}$ to optimize directly for $u$ without the use of any radio data to show that the corruptions themselves are indeed random and not learnable.

\paragraph{RF Baseline}
To reconstruct the footprint of a building in an environment, ~\cite{reconstructioniwcmc} introduces the CLIP+UPerNet network that uses RF information exclusively through radio ray maps, without including any kind of building information during training. The overall architecture of this baseline is illustrated in \Cref{fig:clip-upernet}, where radio ray maps are encoded by a CLIP image encoder and decoded into a building-footprint prediction using UPerNet. Notably, the study employed the WAIR-D dataset with a smaller test set.
In the present work, we replicate the CLIP+UPerNet network, while using it in the current split setting.

\subsection{All-is-building baseline}
\label{sec:all-is-building-baseline}
For evaluation on the real Oslo dataset, we include an \emph{all-is-building} baseline. This is a zero-modality predictor: it does not consume the corrupted map or any RF measurements and instead labels every spatial location as building. We use it because the Oslo crops are dominated by building coverage, so a trivial constant prediction of ``all building'' yields non-trivial IoU and serves as a simple sanity check against which RF-only and fusion models can be compared.

\subsection{Non-AI Fusion Baseline}  \label{sec::bas}

To establish a lower bound for map refinement with both modalities and  without relying on deep learning, we design a non-AI fusion baseline that utilizes the information from all 150 UE--BS pairs in an environment. Our approach begins by classifying each UE--BS pair as either \textit{line-of-sight} (LOS) or \textit{non-line-of-sight} (NLOS). Here, LOS denotes that the direct path between the UE and BS is unobstructed by any building, while NLOS indicates that a building lies along the direct path. These LOS/NLOS flags supplement the information contained in the corrupted map $\tilde{u}$ and provide additional cues about the true building layout.

Empirical observations reveal that the relationship between the UE--BS distance $x$ and the measured path loss is approximately logarithmic. Accordingly, for each of the five frequency bands $$\{2.6, 6, 28, 60, 100\}\,\mathrm{GHz}$$, we model the expected path loss using logarithmic function. In practice we fit $y = a + \log(x)$ via linear least squares regression.
For a given test pair, if the measured path loss falls below $y + \delta$ (with $\delta$ scaling linearly as $0.005x$), the pair is classified as LOS; otherwise, it is marked as NLOS in a given band. A majority vote across the five bands determines the final LOS/NLOS label for a pair.

Once these flags are obtained, we enforce consistency between the LOS/NLOS classification and the building locations in $\tilde{u}$ via a path-based refinement procedure (see \Cref{alg:refinement}). Specifically, at line 2 the algorithm computes the initial violations, where $V_{\text{nlos}}$ denotes NLOS pairs that do not intersect any building, and $V_{\text{los}}$ denotes LOS pairs that erroneously intersect a building. Then, in each iteration (line 4), if $V_{\text{nlos}}$ contains no violations the process terminates (line 5). Otherwise, candidate locations are randomly sampled from the violation points (line 7), and for each candidate location a random building is generated (line 10) and merged with the current map (line 11). If the new map reduces the total number of violations while keeping any increase in $V_{\text{los}}$ bounded (line 13), it is accepted as a candidate improvement. Finally, when the improvement list is not empty we sort it based on total violations first and by $V_{\text{los}}$  if totals are equal (line 20) , the best candidate is then selected to update the refined map $\hat{u}$ (line 21), and the procedure repeats until convergence.

\section{Results and Discussion} \label{sec::res}

\begin{figure*}[h]
    \centering
    \begin{overpic}[width=1\linewidth,trim={10mm 0mm 20mm 0mm},clip]{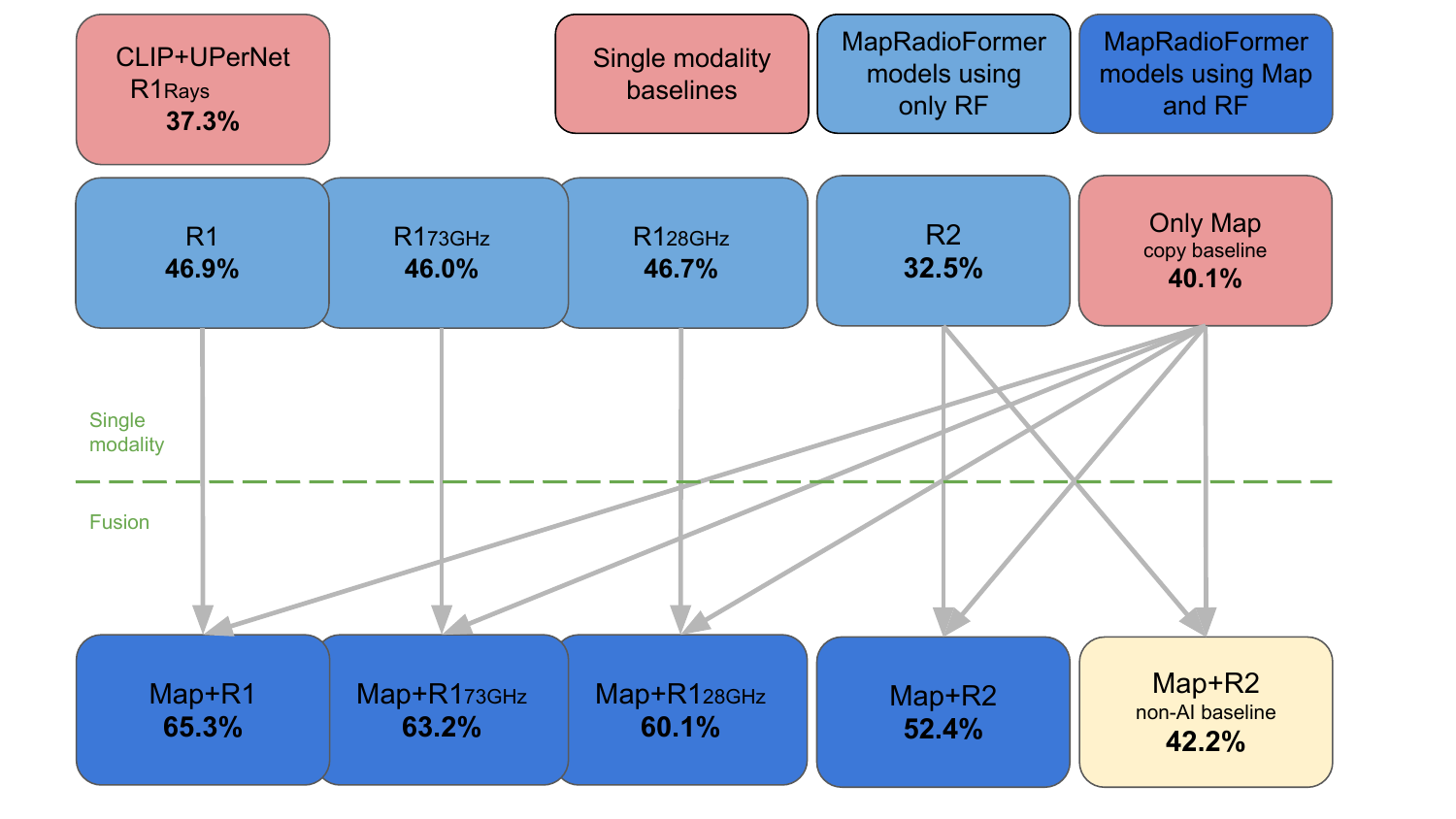}
        \put(13,56.25){\cite{reconstructioniwcmc}} 
    \end{overpic}
    \caption{Visual comparison of the results. Each bloack represents a model from \Cref{tab:model-comparison}. Methods above the dashed line use a single input modality (either the corrupted map or RF data alone), while methods below the dashed line fuse both modalities. Along the horizontal axis, RF granularity decreases from left to right as $\mathrm{R1} \rightarrow \mathrm{R1_{73\,GHz}} \rightarrow \mathrm{R1_{28\,GHz}} \rightarrow \mathrm{R2}$. Vertical arrows link each fusion model to its single-modality counterpart and visualize the IoU gain obtained by fusion. For clarity, models trained without UE-BS subsampling augmentation and the map-only MapRadioFormer are omitted from this plot.}
    \label{fig:results}
\end{figure*}

This section evaluates MapRadioFormer and all baselines on the fixed $1{,}000$-environment WAIR-D test set and then reports additional results on the real Oslo dataset. We present the main quantitative comparison (Macro/Micro IoU, Hausdorff, Chamfer) in \Cref{tab:model-comparison}. Next, we provide qualitative visualizations (\Cref{fig:results,fig:comparative_viz}), and large-area deployment (\Cref{sec:deployment-large}), and a group-wise analysis by corruption factors (\Cref{sec::grev}), robustness to training-time corruption severity, as defined in \Cref{sec:corruption}.

\begin{table*}[ht!]
    \centering
    \caption{Performance evaluation on the WAIR-D test set. Rows are grouped into three categories: (i)~non-learning baselines (the copy baseline that outputs the corrupted map directly and the non-AI fusion baseline of \Cref{sec::bas}), (ii)~learning-based single-modality approaches (CLIP+UPerNet of~\cite{reconstructioniwcmc}~and MapRadioFormer with only map or only RF inputs), and (iii)~fusion-based learning models (MapRadioFormer trained on both modalities). The ``Map'' column indicates whether the corrupted map $\tilde u$ is provided as input; the ``RF'' column specifies the RF granularity (R1 path-level AoA/AoD/ToA, optionally with added noise corresponding to 28 or 73~GHz carrier frequnecies; R2 multi-band path loss; or ray maps as used by CLIP+UPerNet); and the ``Subset'' column indicates whether UE-BS pair subsampling augmentation of \cref{sec:aug} is applied during training. Each configuration is evaluated with Macro IoU, Micro IoU, Hausdorff distance and Chamfer distance (all defined in \Cref{sec:eval-metrics}). The overall best result in each column is \textcolor{blue}{highlighted}.}
    \label{tab:model-comparison}
    \begin{tabular}
    {%
        c|c|c|c||c|c|c|c
    }
    \hline
    \textbf{Method} & \textbf{Map} & \textbf{RF} & \textbf{Subset} & \textbf{Macro IoU} & \textbf{Micro IoU} & \textbf{Hausdorff} & \textbf{Chamfer} \\
    \hline
    \multicolumn{8}{c}{\textbf{Non-Learning Baselines}} \\
    \hline
    Copy Baseline 
      & \cmark
      & \xmark
      & --- 
      & 40.1\% 
      & 39.9\%
      & 151.6m
      & 50.2m
      \\
    Non-AI Baseline 
      & \cmark
      & $R2$
      & \xmark 
      & 42.2\% 
      & 42.0\% 
      & 99.2m
      & 17.0m
      \\
    \hline
    \multicolumn{8}{c}{\textbf{Learning-Based Single-Modality Approaches}} \\
    \hline
    CLIP+UPerNet \cite{reconstructioniwcmc}
      & \xmark
      & $R1_{\mathrm{Rays}}$
      & \cmark
      & 37.3\% 
      & 40.1\%
      & 60.6m
      & 7.0m
      \\
    \hdashline
    \multirow{7}{*}{\centering MapRadioFormer}
     & \cmark
      & \xmark
      & --- 
      & 39.8\% 
      & 40.1\%
      & 131.2m
      & 45.8m
      \\
      & \xmark
      & $R1$
      & \cmark
      & 46.9\% 
      & 49.0\% 
      & 45.8m
      & 4.0m
      \\
    & \xmark
      & $R1$
      & \xmark
      & 45.4\% 
      & 47.7\% 
      & 48.2m
      & 4.2m
      \\
    & \xmark
      & $R1_{73\mathrm{Ghz}}$
      & \cmark
      & 46.0\% 
      & 48.1\% 
      & 47.7m
      & 4.2m
      \\
    & \xmark 
      & $R1_{28\mathrm{Ghz}}$
      & \cmark
      & 46.7\% 
      & 48.7\% 
      & 48.5m
      & 4.1m
      \\
    & \xmark
      & $R2$ 
      & \cmark
      & 32.5\%
      & 36.2\% 
      & 80.6m
      & 10.8m
      \\
    & \xmark
      & $R2$
      & \xmark
      & 31.1\%
      & 35.1\% 
      & 87.0m
      & 15.7m
      \\
    \hline
    \multicolumn{8}{c}{\textbf{Fusion-Based Learning Models}} \\
    \hline
    \multirow{6}{*}{\centering MapRadioFormer}
      & \cmark
      & $R1$
      & \cmark
      & \textbf{\textcolor{blue}{65.3\%}}
      & \textbf{\textcolor{blue}{65.8\%}} 
      & \textbf{\textcolor{blue}{33.9m}}
      & \textbf{\textcolor{blue}{2.0m}}
      \\
    & \cmark
      & $R1$
      & \xmark
      & 59.6\% 
      & 60.8\% 
      & 44.9m
      & 3.2m
      \\
    & \cmark
      & $R1_{73\mathrm{Ghz}}$
      & \cmark
      & 63.2\% 
      & 64.2\% 
      & 39.0m
      & 2.5m
      \\
    & \cmark
      & $R1_{28\mathrm{Ghz}}$
      & \cmark
      & 60.1\% 
      & 61.5\% 
      & 44.0m
      & 3.1m
      \\
    & \cmark 
      & $R2$
      & \cmark
      & 52.4\% 
      & 54.5\% 
      & 67.8m
      & 6.8m
      \\
    & \cmark 
      & $R2$
      & \xmark
      & 55.7\% 
      & 57.8\% 
      & 62.0m
      & 6.1m
      \\
    \hline
    \end{tabular}
\end{table*}

\begin{table*}[ht!]
    \centering
    \small
    \caption{Performance evaluation of fusion-based MapRadioFormer models on the real-world Oslo test set. All rows use the corrupted Oslo map together with RF measurements as input. The ``RF'' column indicates the RF granularity ($R2$ replicated from the WAIR-D path-loss representation, or $R2_{\mathrm{rich}}$ using the full set of real-world KPIs defined in \Cref{sec:granularity-r2rich}); the ``Pre-trained'' column indicates whether the model was first pre-trained on the synthetic WAIR-D dataset; and the ``Trained on Oslo'' column indicates whether the model was (fine-)tuned on Oslo data. All methods are evaluated with Macro IoU, Micro IoU, Hausdorff distance and Chamfer distance, and the overall best result in each column is \textcolor{blue}{highlighted}.}
    \label{tab:model-comparison-oslo}
    \begin{tabular}
    {
        c|c|c||c|c|c|c
    }
    \hline
    \textbf{RF} & \textbf{\shortstack{Pre-trained\\on WAIR-D}} & \textbf{\shortstack{Trained\\on Oslo}} & \textbf{Macro IoU} & \textbf{Micro IoU} & \textbf{Hausdorff} & \textbf{Chamfer} \\
    \hline
    R2
      & \cmark
      & \xmark
      & 40.1\%
      & 39.9\%
      & 47.6
      & 4.0
      \\
    R2
      & \cmark
      & \cmark
      & 60.5\%
      & 60.9\%
      & 31.0
      & 1.8
      \\
    R2
      & \xmark
      & \cmark
      & 62.8\%
      & 63.0\%
      & 31.4
      & 1.7
      \\
    $R2_{\mathrm{rich}}$
      & \xmark
      & \cmark
      & \textbf{\textcolor{blue}{64.9\%}}
      & \textbf{\textcolor{blue}{65.1\%}}
      & \textbf{\textcolor{blue}{27.3}}
      & \textbf{\textcolor{blue}{1.2}}
      \\
    \hline
    \end{tabular}
\end{table*}

\definecolor{lowblue}{RGB}{255,255,255}
\definecolor{highblue}{RGB}{55,55,255}
\definecolor{lowgreen}{RGB}{230,255,230}
\definecolor{highgreen}{RGB}{0,255,0}

\newcommand{\scaledpercentage}[1]{%
  \pgfmathparse{#1-8.9}%
  \xdef\myscaled{\pgfmathresult}%
}

\newcommand{\colorcontrol}[1]{%
  \scaledpercentage{#1}%
  \cellcolor{highblue!\myscaled!lowblue}{#1}\%
}

\newcommand{\colorMacroIoU}[1]{\colorcontrol{#1}}
\newcommand{\colorMicroIoU}[1]{\colorcontrol{#1}}

\begin{table*}[h]
    \setlength{\tabcolsep}{5pt}
    \begin{tabular}{c|c|c|c||c c|c c}
        \toprule
        \multirow{2}{*}{Shift Range} & \multirow{2}{*}{Simplify} & \multirow{2}{*}{Removal Percentage} & \multirow{2}{*}{Count} &  \multicolumn{2}{c|}{Corrupted Map IoU} & \multicolumn{2}{c}{R1 Fusion IoU}  \\
        & & & & Macro & Micro & Macro & Micro \\
        \midrule
        $s \leq 1.46m$ & \xmark  & $r \leq 46.15\%$           & 99 & \colorMacroIoU{64.8} & \colorMicroIoU{64.0} & \colorMacroIoU{77.2} & \colorMicroIoU{76.9} \\
        $s \leq 1.46m$ & \cmark & $r \leq 46.15\%$           & 41 & \colorMacroIoU{64.3} & \colorMicroIoU{66.6} & \colorMacroIoU{73.1} & \colorMicroIoU{74.4} \\
        $s \leq 1.46m$ & \xmark  & $46.15\% < r \leq 57.14\%$ & 114 & \colorMacroIoU{45.8} & \colorMicroIoU{44.8} & \colorMacroIoU{70.5} & \colorMicroIoU{71.2} \\
        $s \leq 1.46m$ & \cmark & $46.15\% < r \leq 57.14\%$ & 49  & \colorMacroIoU{45.2} & \colorMicroIoU{45.8} & \colorMacroIoU{66.2} & \colorMicroIoU{66.9} \\
        $s \leq 1.46m$ & \xmark  & $57.14\% < r \leq 69.31\%$ & 87  & \colorMacroIoU{40.2} & \colorMicroIoU{44.2} & \colorMacroIoU{61.3} & \colorMicroIoU{63.5} \\
        $s \leq 1.46m$ & \cmark & $57.14\% < r \leq 69.31\%$ & 32  & \colorMacroIoU{28.5} & \colorMicroIoU{29.5} & \colorMacroIoU{53.6} & \colorMicroIoU{55.8} \\
        $s \leq 1.46m$ & \xmark  & $r > 69.31\%$              & 107 & \colorMacroIoU{15.2} & \colorMicroIoU{13.9} & \colorMacroIoU{57.6} & \colorMicroIoU{58.6} \\
        $s \leq 1.46m$ & \cmark & $r > 69.31\%$              & 32  & \colorMacroIoU{16.0} & \colorMicroIoU{14.8} & \colorMacroIoU{58.0} & \colorMicroIoU{58.4} \\
        \midrule
        $1.46m < s \leq 2.5m$ & \xmark  & $r \leq 46.15\%$           & 45 & \colorMacroIoU{64.9} & \colorMicroIoU{62.0} & \colorMacroIoU{78.7} & \colorMicroIoU{78.9} \\
        $1.46m < s \leq 2.5m$ & \cmark & $r \leq 46.15\%$           & 19 & \colorMacroIoU{57.4} & \colorMicroIoU{58.7} & \colorMacroIoU{71.3} & \colorMicroIoU{72.6} \\
        $1.46m < s \leq 2.5m$ & \xmark  & $46.15\% < r \leq 57.14\%$ & 30 & \colorMacroIoU{43.8} & \colorMicroIoU{40.2} & \colorMacroIoU{67.8} & \colorMicroIoU{68.8} \\
        $1.46m < s \leq 2.5m$ & \cmark & $46.15\% < r \leq 57.14\%$ & 14 & \colorMacroIoU{42.4} & \colorMicroIoU{44.3} & \colorMacroIoU{62.8} & \colorMicroIoU{63.4} \\
        $1.46m < s \leq 2.5m$ & \xmark  & $57.14\% < r \leq 69.31\%$ & 45 & \colorMacroIoU{43.4} & \colorMicroIoU{42.6} & \colorMacroIoU{65.4} & \colorMicroIoU{66.0} \\
        $1.46m < s \leq 2.5m$ & \cmark & $57.14\% < r \leq 69.31\%$ & 18 & \colorMacroIoU{30.4} & \colorMicroIoU{31.5} & \colorMacroIoU{64.9} & \colorMicroIoU{65.5} \\
        $1.46m < s \leq 2.5m$ & \xmark  & $r > 69.31\%$              & 45 & \colorMacroIoU{15.2} & \colorMicroIoU{16.8} & \colorMacroIoU{58.2} & \colorMicroIoU{58.4} \\
        $1.46m < s \leq 2.5m$ & \cmark & $r > 69.31\%$              & 18 & \colorMacroIoU{18.5} & \colorMicroIoU{20.3} & \colorMacroIoU{54.6} & \colorMicroIoU{53.9} \\
        \midrule
        $s > 2.5m$ & \xmark  & $r \leq 46.15\%$           & 40 & \colorMacroIoU{63.1} & \colorMicroIoU{63.8} & \colorMacroIoU{71.0} & \colorMicroIoU{71.9} \\
        $s > 2.5m$ & \cmark & $r \leq 46.15\%$           & 8  & \colorMacroIoU{60.5} & \colorMicroIoU{63.3} & \colorMacroIoU{74.3} & \colorMicroIoU{74.9} \\
        $s > 2.5m$ & \xmark  & $46.15\% < r \leq 57.14\%$ & 36 & \colorMacroIoU{40.0} & \colorMicroIoU{34.1} & \colorMacroIoU{63.0} & \colorMicroIoU{63.7} \\
        $s > 2.5m$ & \cmark & $46.15\% < r \leq 57.14\%$ & 19 & \colorMacroIoU{39.2} & \colorMicroIoU{42.6} & \colorMacroIoU{57.2} & \colorMicroIoU{59.7} \\
        $s > 2.5m$ & \xmark  & $57.14\% < r \leq 69.31\%$ & 41 & \colorMacroIoU{30.9} & \colorMicroIoU{28.8} & \colorMacroIoU{58.1} & \colorMicroIoU{59.2} \\
        $s > 2.5m$ & \cmark & $57.14\% < r \leq 69.31\%$ & 13 & \colorMacroIoU{35.1} & \colorMicroIoU{33.4} & \colorMacroIoU{58.5} & \colorMicroIoU{62.4} \\
        $s > 2.5m$ & \xmark  & $r > 69.31\%$              & 38 & \colorMacroIoU{12.0} & \colorMicroIoU{8.9}  & \colorMacroIoU{58.5} & \colorMicroIoU{60.1} \\
        $s > 2.5m$ & \cmark & $r > 69.31\%$              & 10 & \colorMacroIoU{27.5} & \colorMicroIoU{36.5} & \colorMacroIoU{61.2} & \colorMicroIoU{59.8} \\
        \midrule
        \bottomrule
    \end{tabular}
\caption{Detailed evaluation for input maps and the best fusion model. For each combination of corruption parameters the table reports Macro IoU for (i)~using the corrupted input map $\tilde u$ directly as a prediction (copy baseline) and (ii)~the best fusion model $\mathrm{MapRadioFormer_{Map+R1}}$. Rows group the test environments by positional shift range (using the percentile bins of 50\% and 80\% from \Cref{tab:building-shift}), whether polygon simplification was applied, and by the per-environment building removal rate (using quartile-style bins derived from \Cref{fig:removal-dist}). This arrangement makes it possible to identify which corruption regimes are most challenging and to verify that the fusion model consistently improves over the corrupted input across all corruption regimes. Cell shading encodes the Macro IoU value on a common color scale to aid visual comparison.}
\label{tab:detailed_eval}
\end{table*}

\begin{table}[ht!]
    \setlength{\tabcolsep}{5pt}
    \centering
    \caption{The results of severity experiments for $R1$ and $R2$ levels of granularity, reported as Macro and Micro IoU on the WAIR-D validation set. At Level~1 the corruption parameters used during training match those used at test time (building removal rate $57\%$, shift distribution from \Cref{tab:building-shift}, simplification rate $25\%$). Levels~1.5 and 2 correspond to proportionally more severe corruption during training (building removal rates $71.3\%$ and $78.5\%$, shift values scaled by $1.5\times$ and $2\times$, and simplification rates $37.5\%$ and $50\%$, respectively). The table shows that simply training under harsher corruption does not improve validation performance, and motivates using Level~1 severity for all experiments.}
    \label{tab:severity}
    \begin{tabular}
    {
        c|c|c|c
    }
    \hline
    \textbf{Granularity} & \textbf{Severity} & \textbf{Macro IoU} & \textbf{Micro IoU} \\
    \hline
    \multirow{3}{*}{R1}
     & Level 1   & 65.8\% & 66.2\%\\
     & Level 1.5 & 65.4\% & 65.9\%\\
     & Level 2   & 64.1\% & 64.8\%\\
     
    \hline
    \multirow{3}{*}{R2}
     & Level 1   & 56.0\% & 58.2\%\\
     & Level 1.5 & 50.5\% & 52.9\%\\
     & Level 2   & 51.0\% & 53.5\%
    \end{tabular}
\end{table}

\begin{table}[ht!]
    \setlength{\tabcolsep}{5pt}
    \centering
    \caption{Ablation study results for the full R1 configuration versus versions with AoA or AoD removed. The reference ``R1'' row corresponds to the best $\mathrm{MapRadioFormer_{Map+R1}}$ model trained with the complete per-path AoA, AoD, and ToA features (as defined in \Cref{sec:granularity-angles}); the ``No AoA'' and ``No AoD'' rows use the same architecture and training setup but with the AoA or AoD zeroed out from the feature vector. Macro and Micro IoU are reported on the WAIR-D test set, enabling a direct assessment of the individual contribution of AoA and AoD to the fusion model.}
    \label{tab:ablation}
    \begin{tabular}{c|c|c}
        \hline
        \textbf{Configuration} & \textbf{Macro IoU} & \textbf{Micro IoU} \\
        \hline
        R1 & 65.3\% & 65.8\% \\
        No AoA    & 62.5\% & 63.3\% \\
        No AoD    & 61.6\% & 62.6\% \\
        \hline
    \end{tabular}
\end{table}

\noindent\textbf{Overview of comparisons.}
\Cref{tab:model-comparison} groups methods into (i) non-learning baselines, (ii) single-modality learning (map-only or RF-only), and (iii) fusion models. Each method is characterized by its input modalities:
\begin{itemize}
    \item \textbf{Map} column indicates whether the corrupted map obtained by the procedure described in \cref{sec:corruption} is used as input to the algorithm.
    \item \textbf{RF} column shows the granularity of RF data and the added level of noise, if applicable.
    \item \textbf{Subset} column refers to UE-BS pair subsampling augmentation described in \cref{sec:aug}.
\end{itemize}
Each method is evaluated using both Macro and Micro IoU metrics, as well as Hausdorff \cite{hausdorff_grundzuge_1914} and Chamfer \cite{10.5555/1622943.1622971} distances, all of which are formally defined in \Cref{sec:eval-metrics}.

Our discussion first examines the contributions of individual modalities and their fusion, before delving into the detailed evaluations, experiments under increased corruption severity, and input ablation studies.

\subsection{No radio}
We evaluate two methods that rely exclusively on map tokens. The copy baseline, which directly employs the corrupted map $\tilde{u}$ without refinement, achieves a mean IoU of 40.1\% relative to the ground-truth map $u$. In contrast, a ViT-based model trained solely on $\tilde{u}$ attains an IoU of 39.8\%, indicating that it fails to capture any corruption patterns for recovering accurate building footprints.

\subsection{No map}
To evaluate the standalone contribution of radio information for environment reconstruction, we conducted experiments in which the map modality is omitted, replacing the input with a zero-valued placeholder. As shown in \Cref{tab:model-comparison}, when using the detailed R1 radio tokens, the MapRadioFormer model achieves a Macro IoU of 46.9\% with UE-BS subset selection enabled, which slightly decreases to 45.4\% without it. Under a similar experimental setup, the CLIP+UPerNet network \cite{reconstructioniwcmc}, which leverages ray maps, exhibits substantially lower performance in terms of Macro IoU, achieving a score of 37.3\%. When Gaussian noise is introduced to simulate real-world imperfections in the angle features, the IoU values are 46.0\% and 46.7\% for the two noise configurations. In contrast, employing the coarser R2 radio tokens results in significantly lower performance, with Macro IoU scores of 32.5\% and 31.0\% under analogous conditions. These results highlight that the granularity of the radio features is critical for accurate reconstruction, as the detailed R1 tokens consistently provide more informative structural cues. Among all single-modality approaches, the MapRadioFormer neural network utilizing R1-level RF data achieves the lowest Hausdorff and Chamfer distances, measuring 45.8 meters and 4.0 meters, respectively.

\subsection{Fusion}
Prior to training MapRadioFormer model on map and radio tokens, a non-AI Fusion Baseline was established by combining corrupted maps with radio path information (see \cref{sec::bas}), providing a reference for comparison. When both modalities are integrated, the best model achieves IoU of 65.3\% under the $R1$ radio token configuration with device subsampling enabled. In alternative configurations: using either $R2$ or $R1$ with noisy addition the IoU ranges from 55.7\% to 63.2\%. Compared to their respective RF-only counterparts, fusion models demonstrate superior performace in terms of Hausdorff and Chamfer distances. Specifically, for the R1 level of RF data with device subsampling enabled, the model achieves improvements of approximately 1.35 times in Hausdorff distance and 2 times in Chamfer distance. This substantial improvement over the individual modality baselines demonstrates that the spatial context encoded in radio signals effectively compensates for the systematic errors present in the corrupted maps. 

Additionally, we notice consistency between the individual performances of $R1$ and $R2$ radio granularities and the ones obtained in combination with the corrupted maps. Collectively, these results validate the central hypothesis that integrating detailed radio information with incorrect map inputs enables a more accurate reconstruction of the true environment.

\subsection{Real-world results}
\label{sec:real-world-results}

We evaluate four fusion variant of MapRadioFormer on the real Oslo dataset. We consider the configurations, reported in \Cref{tab:model-comparison-oslo}: (i) a model pre-trained on WAIR-D and evaluated on Oslo without Oslo training; (ii) a model pre-trained on WAIR-D and fine-tuned on Oslo; (iii) a model trained on Oslo from scratch using the $R2$ feature vector replicated from the single aggregated pathloss value available in Oslo; and (iv) a model trained on Oslo from scratch using the full set of real-world $R2_{\mathrm{rich}}$ radio measurements. In this setup, the ratio of real to synthetic data increases progressively across experiments.

The WAIR-D pre-trained model transferred directly to Oslo reaches only 40.1\% Macro IoU, indicating that synthetic pre-training alone does not generalize to Oslo. Fine-tuning on Oslo raises this number to 60.5\%, and training on Oslo from scratch with the replicated $R2$ input yields a further improvement to 62.8\% Macro IoU, indicating that the former model was not able to fully overcome the biases inherited from the synthetic pre-training. Replacing the replicated pathloss value with the full real-world $R2_{\mathrm{rich}}$ measurements while keeping the Oslo-from-scratch training regime gives the best Oslo result across all four metrics: 64.9\% Macro IoU, 65.1\% Micro IoU, Hausdorff distance 27.3, and Chamfer distance 1.2. This trend suggests that the informativeness of the real radio measurements, rather than the amount of synthetic pre-training, is the dominant factor on Oslo.

Compared to our best fusion result on the synthetic WAIR-D test set (65.3\% Macro IoU obtained by MapRadioFormer$_{\mathrm{Map+R1}}$), the best Oslo fusion model is essentially on par at 64.9\% Macro IoU. We interpret this parity cautiously: we have not performed controlled ablations that would isolate the role of the architecture from properties of the data itself, such as the distribution of building footprints, or the geographic coverage of the Oslo test region. The match between the two datasets is therefore best read as preliminary evidence that the proposed fusion model transfers to a real deployment without catastrophic degradation.

\subsection{Visual comparison}

For improved comparative interpretability, we refer the reader to \Cref{fig:results}. Each method in the figure is evaluated in terms of Macro IoU. The methods positioned above the dashed line utilize a single input modality, either RF data or the map alone, whereas the methods below the dashed line integrate both modalities. The granularity of RF data decreases from left to right: $\mathrm{R1} \rightarrow \mathrm{R1_{73GHz}} \rightarrow \mathrm{R1_{28GHz}} \rightarrow \mathrm{R2}$. Arrows in the figure indicate the performance improvements of fusion-based methods over their single-modality counterparts. As observed in the figure, fusion-based approaches consistently achieve superior performance compared to their corresponding single-modality methods. Notably, even when employing the lowest granularity R2 RF data, our MapRadioFormer model, when combined with map information, outperforms all methods above the dashed line. The upper left corner of the figure provides a visual explanation of the color coding. For simplicity, we have omitted methods trained without UE-BS subsampling augmentation and the MapRadioFormer model trained exclusively on the map.

        \begin{figure*}
            \centering
            \includegraphics[width=0.86\linewidth]{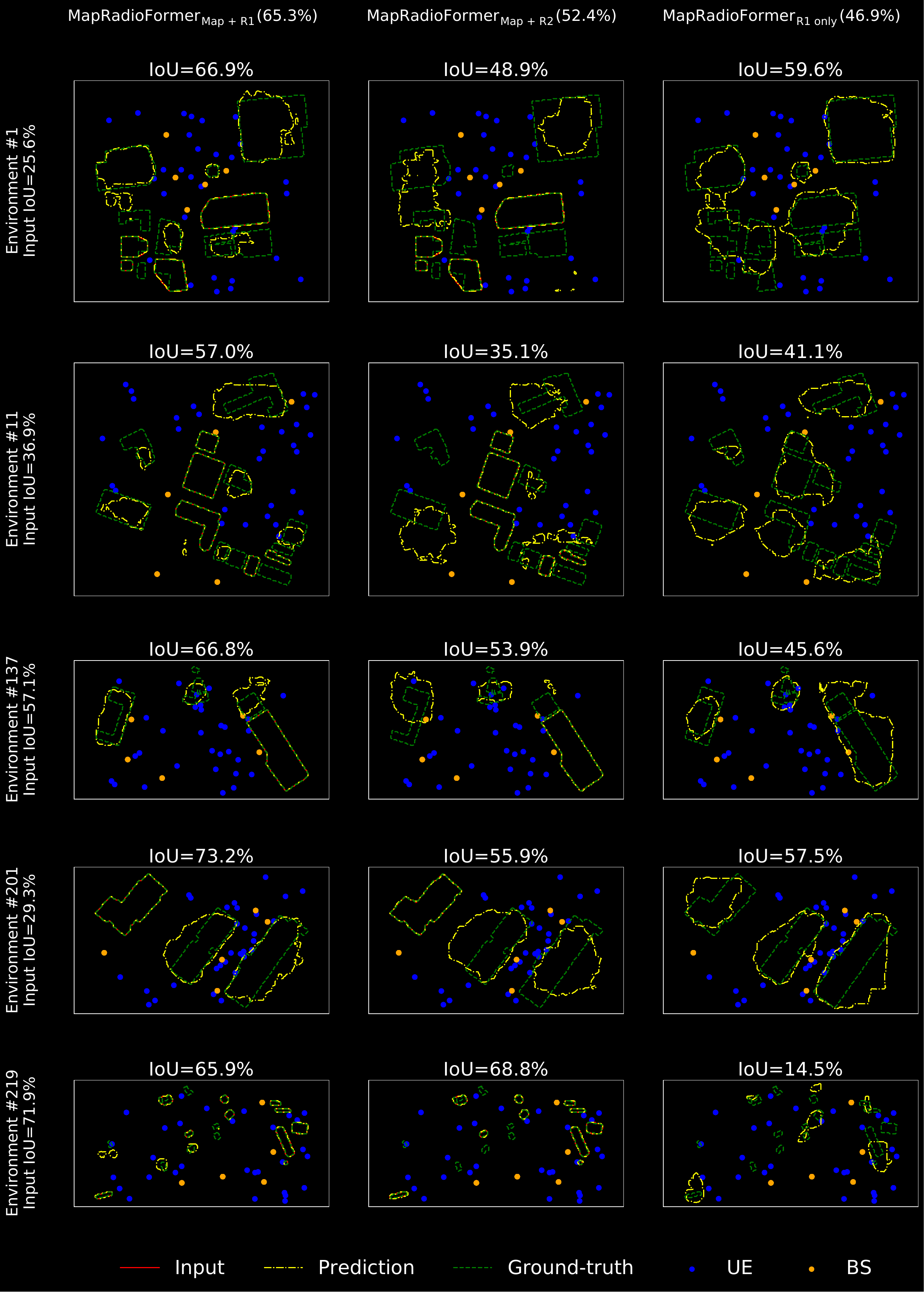}
            \caption{Comparative visualizations for the performance of 3 models on 5 environments. Each row corresponds to one WAIR-D test environment. Each environment is visualized with three subfigures that superimpose the contours of the \colorbox{black}{\textcolor{inputcolor}{inaccurate input map}}, the \colorbox{black}{\textcolor{predcolor}{predicted map}}, and the \colorbox{black}{\textcolor{gtcolor}{ground-truth map}}. The vertical text on the left side of each row shows the IoU of using the corrupted input $\tilde{u}$ directly as prediction (copy baseline), while the three subfigures show predictions of MapRadioFormer trained on: (i)~inaccurate map + R1 RF data, (ii)~inaccurate map + R2 RF data, and (iii)~R1 RF data only. Per-environment IoU values are reported above the corresponding subfigures.}
            \label{fig:comparative_viz}
        \end{figure*}

\subsection{Comparative visualization} 
\Cref{fig:comparative_viz} presents a side‐by‐side comparison of reconstructions generated by the three models across five distinct environments. We visualize the predictions of MapRadioFormer models trained on the: (i) inaccurate map and R1 level of RF data, (ii) inaccurate map and R2 level of RF data, (iii) only R1. At the left side of each row we report the IoU of using $\tilde{u}$ directly as a prediction. Each map visualizes the contours of the \colorbox{black}{\textcolor{inputcolor}{inaccurate input map}}, the \colorbox{black}{\textcolor{predcolor}{predicted map}}, and the \colorbox{black}{\textcolor{gtcolor}{ground-truth map}}. 

In the first four environments, the $\mathrm{MapRadioFormer_{Map+R1}}$ fusion model exhibits superior performance by demonstrating predictions aligned closely with the ground truth, accurately capturing both the overall structure and finer building details. Conversely, the $\mathrm{MapRadioFormer_{Map+R2}}$ approach, tends to recover fewer details, often delineating only the most prominent structures resulting in a modestly lower IoU.
The performance of the only radio model is markedly different. In the first environment, it compensates for the absence of map input by predicting a single, large building, which in some cases yields a competitive IoU relative to the other methods. However, as the environmental conditions vary, this model exhibits a more exploratory behavior. In the final environment, for instance, the approach using only RF produces fragmented reconstructions that fail to capture the spatial details.
Notably, in the last environment, the fusion model with R1 loses its IoU advantage. Although it predicts a greater number of smaller structures corresponding to buildings present in the ground truth, the looser boundaries and slight misalignments of these additional predictions diminish the overall metric relative to the Map+R2 configuration.

\subsection{Detailed evaluation} \label{sec::grev}
In order to assess the impact of different corruption levels on model's performance, \Cref{tab:detailed_eval} presents a detailed analysis of the metrics achieved by two models: (i) using corrupted input map input $\tilde{u}$ directly as a prediction and (ii) our best performing fusion model $\mathrm{MapRadioFormer_{Map+R1}}$. The results are partitioned by positional shift range, building removal percentile, and polygon simplification. Specifically, the aim is to identify which corruption groups are more challenging and whether the group-wise performance trends are consistent between the two considered models. Across all shift ranges, the outputs of the fusion model consistently outperform the inaccurate inputs. For example, under minimal corruption  $s \leq 1.46\,\mathrm{m}$ and $r \leq 46.15\%$, baseline Macro IoUs are  64.8\% (without simplification) and 64.3\% (with simplification), while fusion improves these to 77.2\% and 73.1\%, respectively. As removal rates increase, baseline performance degrades markedly, for $r > 69.31\%$, whereas fusion maintains significantly higher scores. Similar patterns emerge for the intermediate $1.46\,\mathrm{m} < s \leq 2.5\,\mathrm{m}$ and high $s > 2.5\,\mathrm{m}$ shift ranges. The results indicate that the performance of the models does not exhibit a consistent pattern with respect to polygon simplification. For enhanced interpretability, we color code the results: The darker shades indicate a better performance, and the less saturated colors indicate lower IoU.

\subsection{Severity experiments}
\label{sec:res-severity}

As shown in \Cref{tab:severity}, increasing the severity of corruption during training does not lead to a substantial improvement in performance on the validation set for either R1 or R2 granularity levels. Based on the results obtained, we select the optimal severity level for subsequent experiments and accordingly present the results on the validation set. From the table, it can be observed that Level 1 corruption severity yields the best performance.

\subsection{Ablations}
\label{sec:ablations}

In order to assess the individual contributions of the angular features within our $R1$ radio representation, we perform ablation studies by removing either the AoA or AoD from the full 19-dimensional vector. Recall that for each UE--BS pair, up to five propagation paths are considered; omitting either one of the angular feature reduces dimensionality to 14.

\Cref{tab:ablation} reports the performance of the $\mathrm{MapRadioFormer_{Map+R1}}$ model and its ablated variants. Excluding AoA results in a Macro IoU of 62.5\% and a Micro IoU of 63.3\%, while omitting AoD yields 61.6\% and 62.6\%, compared to 65.3\% and 65.8\% for the original configuration. These outcomes suggest that while AoA and AoD add complementary value, their combination can achieve higher performance.

\subsection{Scalability}
\label{sec:scale}

\begin{figure}
    \centering
    \includegraphics[width=\linewidth, trim ={5mm 0mm, 5mm, 0mm}, clip]{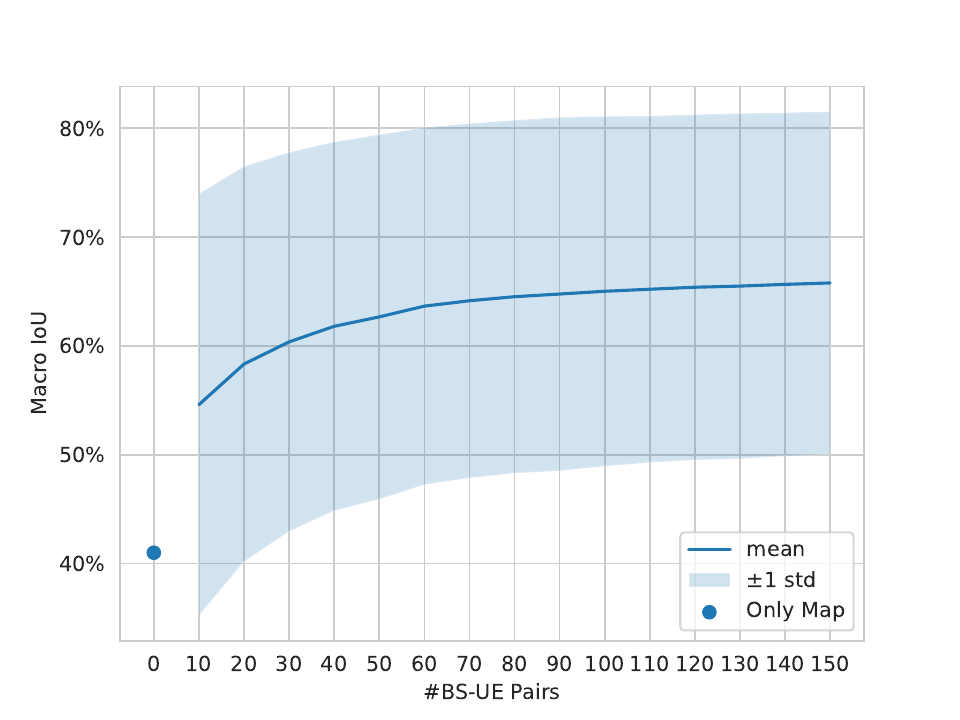}
    \caption{Performance of the model with increasing number of BS-UE pairs. The curve reports the Macro IoU of the best fusion model $\mathrm{MapRadioFormer_{Map+R1}}$ on the WAIR-D test set as the number of available BS-UE pairs is increased from $10$ to $150$ in steps of $10$, where each successive step is a strict superset of the previous one. The shaded band reflects the standard deviation of IoU. The copy baseline is shown with a dot as a 0 BS-UE pair configuration for reference.}
    \label{fig:scale}
\end{figure}

To evaluate the scalability of the proposed framework, we perform an experiment in which the number of available BS–UE pairs is gradually increased from 10 to 150 in increments of 10, as illustrated in \Cref{fig:scale}. At each step, the newly considered set of pairs forms a strict superset of the preceding configuration. The results indicate a marked improvement in reconstruction quality when incorporating even a limited number of radio measurements. Specifically, the use of only 10 BS–UE pairs already yields an improvement of over 13\% compared to the map-only baseline. The performance improvement exhibits a logarithmic growth trend with respect to the number of added pairs, attaining 65.8\% IoU when all 150 BS–UE pairs are employed, compared to 54.6\% when only 10 pairs are utilized. An increase in the number of pairs is also accompanied by a reduction in variability, with the standard deviation decreasing from 19.4\% to 15.7\%.

\subsection{Mobility}
\label{sec:mobile}

\begin{figure}[t]
    \centering
    \includegraphics[width=\linewidth]{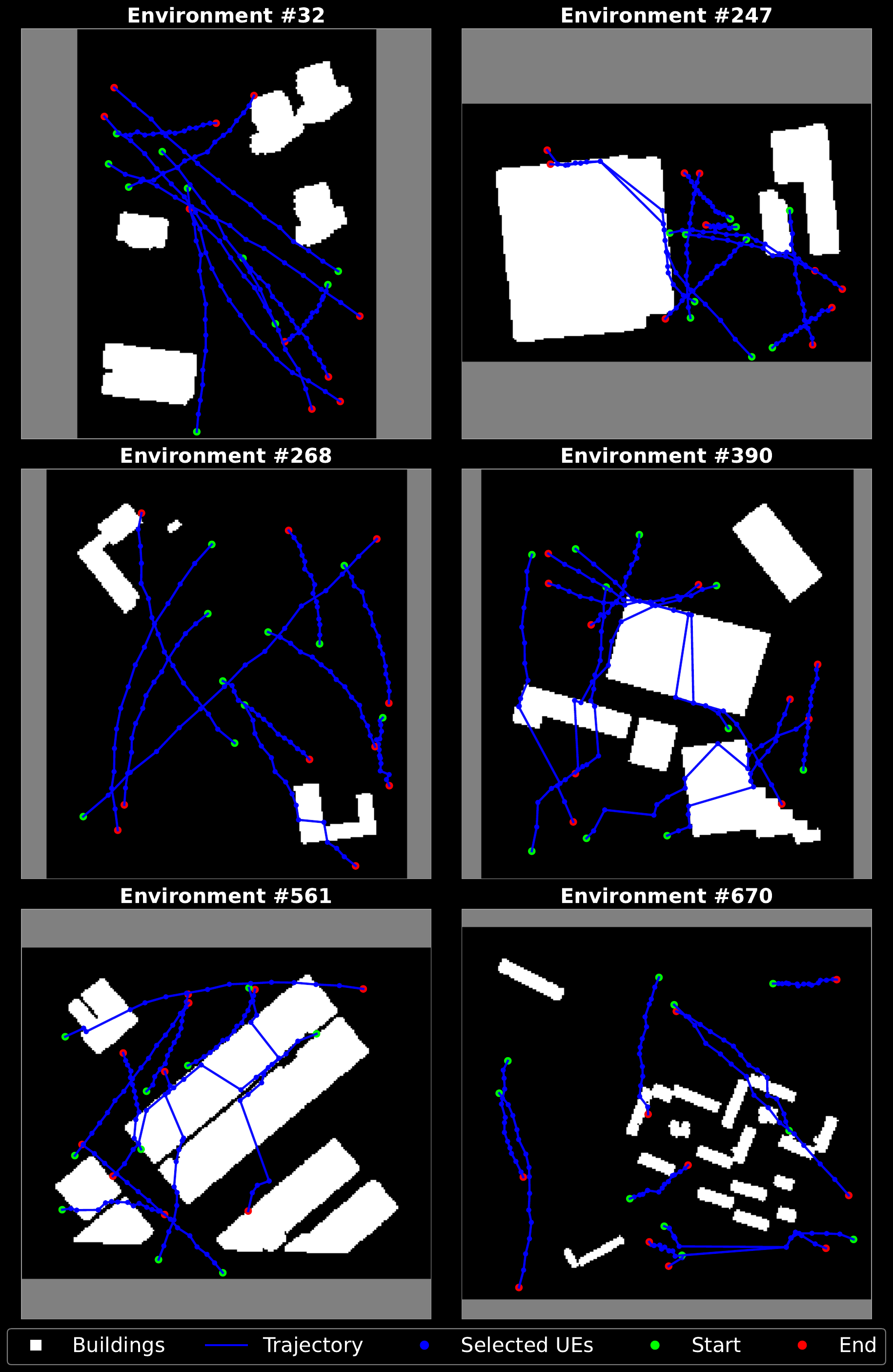}
    \caption{Generated mobility trajectories for six random environments in Scenario~2. Each environment contains 10 UE trajectories, each defined between a starting and an ending position and sampled over 15 timeframes along a quadratic Bezier curve. Gray borders indicate padding.}
    \label{fig:mobile-trajectory-grid}
\end{figure}

\begin{figure*}[t]
    \centering
    \begin{subfigure}[t]{0.32\textwidth}
        \centering
        \includegraphics[width=\linewidth]{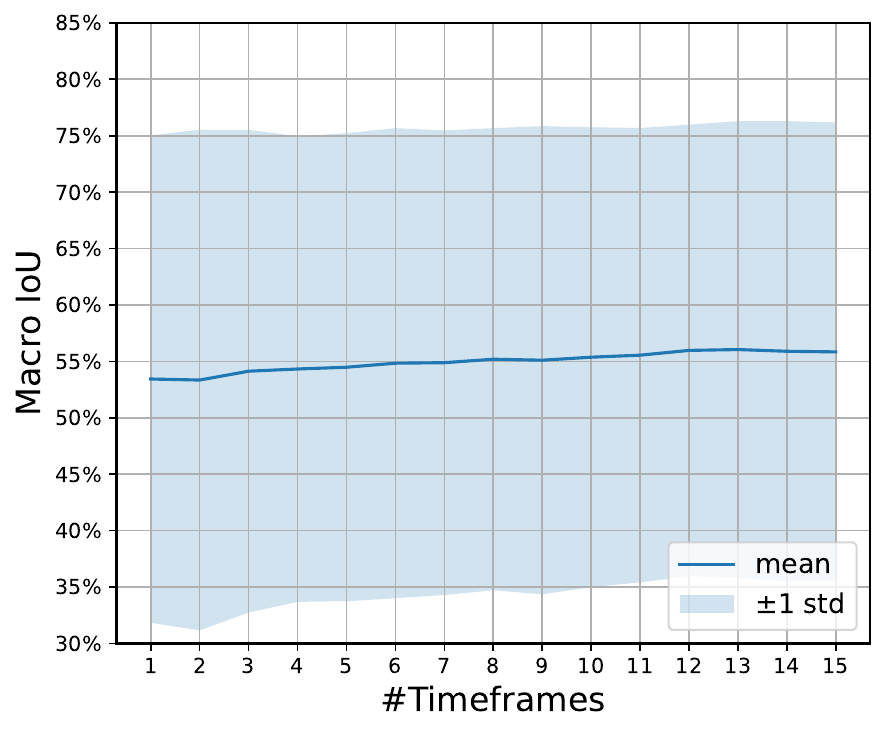}
        \caption{Trajectory-based MapRadioFormer aggregation.}
        \label{fig:mobile-trajectory}
    \end{subfigure}
    \hfill
    \begin{subfigure}[t]{0.32\textwidth}
        \centering
        \includegraphics[width=\linewidth]{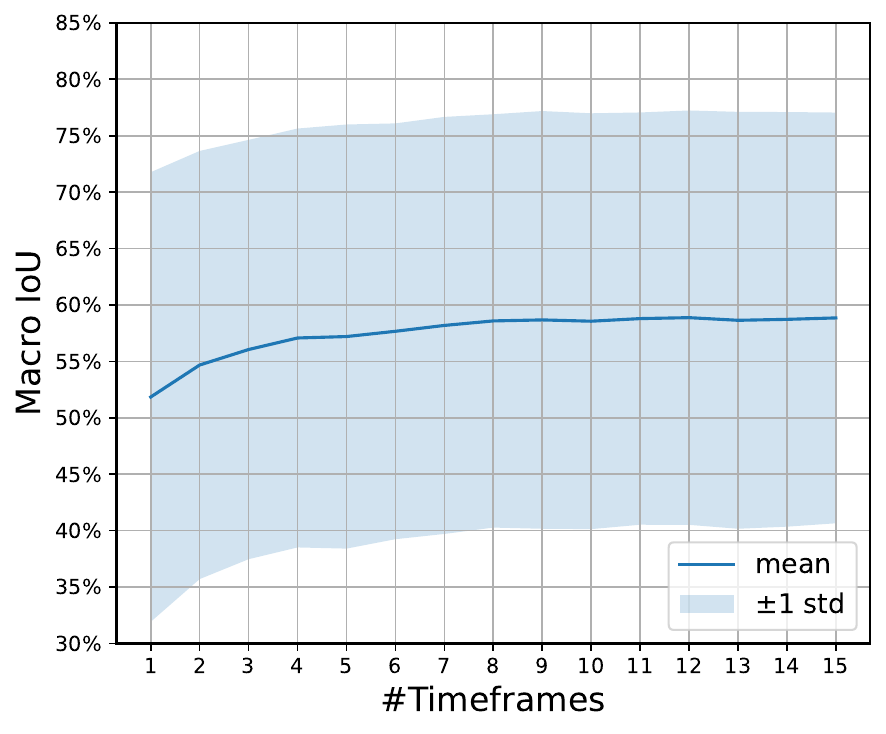}
        \caption{Random MapRadioFormer aggregation.}
        \label{fig:mobile-random}
    \end{subfigure}
    \hfill
    \begin{subfigure}[t]{0.32\textwidth}
        \centering
        \includegraphics[width=\linewidth]{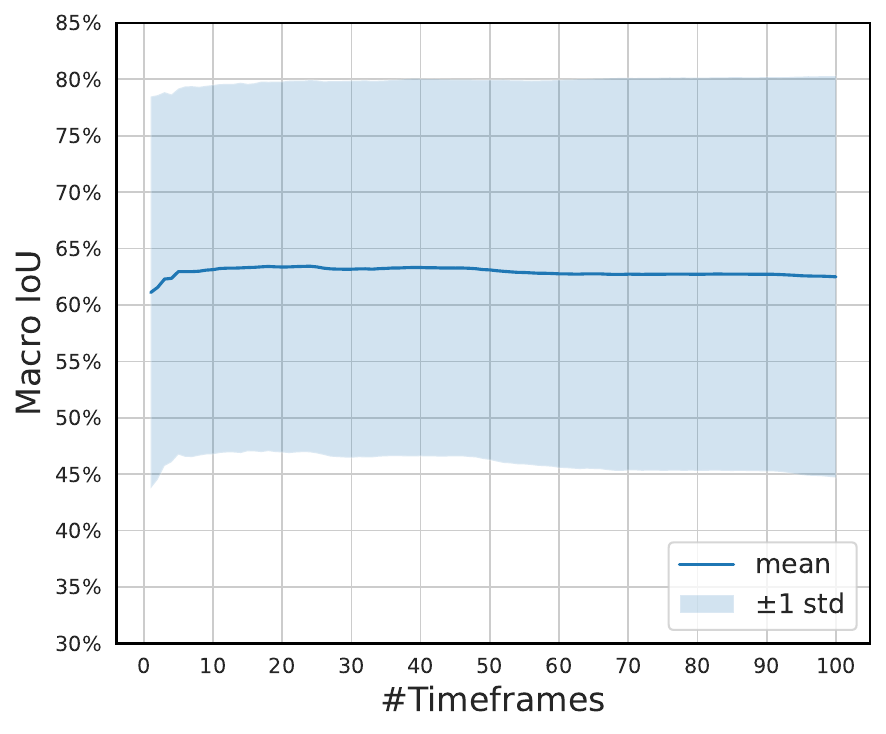}
        \caption{Frame-wise averaging aggregation.}
        \label{fig:mobile-averaging}
    \end{subfigure}
    \caption{Comparison of three mobility aggregation strategies.  \Cref{fig:mobile-trajectory} shows the trajectory-based experiment cummulatively aggregated by MapRadioFormer model. \Cref{fig:mobile-random} shows the random-sampling experiment cummulatively aggregated by MapRadioFormer model. \Cref{fig:mobile-averaging} shows the aggregation-by-averaging experiment.}
    \label{fig:mobile}
\end{figure*}

For the mobility setting, we utilize Scenario~2 of the WAIR-D dataset, which comprises a single BS and 10,000 UEs distributed across 100 environments. In this scenario, the term ``mobility'' reflects that the UEs are not fixed at one location but are spread across different positions over time, mimicking movement through the environment.

\subsubsection{MapRadioFormer aggregation}
To model consistent UE motion, we construct 10 mobility trajectories per environment. Specifically, we select 10 UEs as starting points and 10 UEs as ending points. For each start--end pair, we denote the start location by $P_s$ and the end location by $P_e$, and compute the midpoint $M = \frac{P_s + P_e}{2}$. This midpoint serves as the control point of a 3-point quadratic Bezier curve, ensuring that the generated path remains centered between the chosen endpoints while still following a smooth non-linear trajectory rather than a straight segment. The resulting curve is defined as 
$$B(t) = (1-t)^2 P_s + 2(1-t)t M + t^2 P_e,\; t \in [0,1].$$
We then sample 15 equidistant points along this curve and, for every sampled point, select the geographically closest UE. \Cref{fig:mobile-trajectory-grid} visualizes the generated paths for six random environments.

At the first timeframe, the network is fed with the RF data between the BS and the 10 initial UEs. Subsequently, at each timeframe, the 10 UEs corresponding to the current sampled positions are cumulatively added to the UEs collected from all previous timeframes, and the full accumulated set is fed to the network.

The resulting cumulative trajectory-based aggregation yields a gradual improvement in performance, with IoU increasing from 53.4\% at the first timeframe to 56.1\% at the last timeframe, following an almost monotonic trend, as shown in \Cref{fig:mobile-trajectory}. Standard deviation is around 20-22\%.

For comparison, we also evaluate a cumulative variant in which the UEs added over time are sampled randomly rather than selected to replicate movement along the constructed trajectories. This simulates a scenario where the elasped time between two subsequent frames is much higher. In this case, IoU increases from 51.8\% to 58.9\%, again almost monotonically, as shown in \Cref{fig:mobile-random}. Standard deviation is in 18-20\% range.

Taken together, these cumulative experiments indicate that progressively adding fresh UE observations is beneficial in this simulated setting. The stronger trend obtained with random cumulative sampling further suggests that broader spatial coverage may currently be more useful to the model than preserving a trajectory-consistent motion pattern.

\subsubsection{Averaging aggregation}

But what happens when the model's capacity is reached: either due to memory limitations or due to the fact that model has seen at most 150 pairs during training and can not take advantage of more pairs during inference? In this scenario, rather than relying on MapRadioFormer to do the aggregation, we aggregate the predictions of each timeframe by takeing the per-pixel average. And hence, each network pass is treated as a separate timeframe.

To achieve the goal, we partition UE-BS pairs into 100 temporal frames, each containing 100 pairs. Each time frame simulates a snapshot of activity. The network is applied independently to each frame, and the reconstructed maps are subsequently aggregated by averaging along the temporal dimension.

As illustrated in \Cref{fig:mobile-averaging}, the benefits of combining multiple frames using a manual aggregation strategy are limited. A single frame yields an IoU of 60.8\%, which increases modestly to 63.6\% when aggregating 18 frames, but decreases slightly to 62.6\% when all frames are incorporated. No notable variation in standard deviation is observed with an increasing number of time frames, remaining consistently close to 17\%.

Overall, these findings suggest that in mobile scenarios, limited temporal aggregation enhances robustness by mitigating random variations. Nevertheless, the observed decline in performance with extensive aggregation suggests that the networks's capacity is reached and highlights the need for frameworks specifically tailored to mobile settings in order to efficiently exploit larger numbers of time frames.

\begin{figure*}[p]
    \centering
    \includegraphics[
        width=\textwidth,
        height=0.9\textheight,
        keepaspectratio
    ]{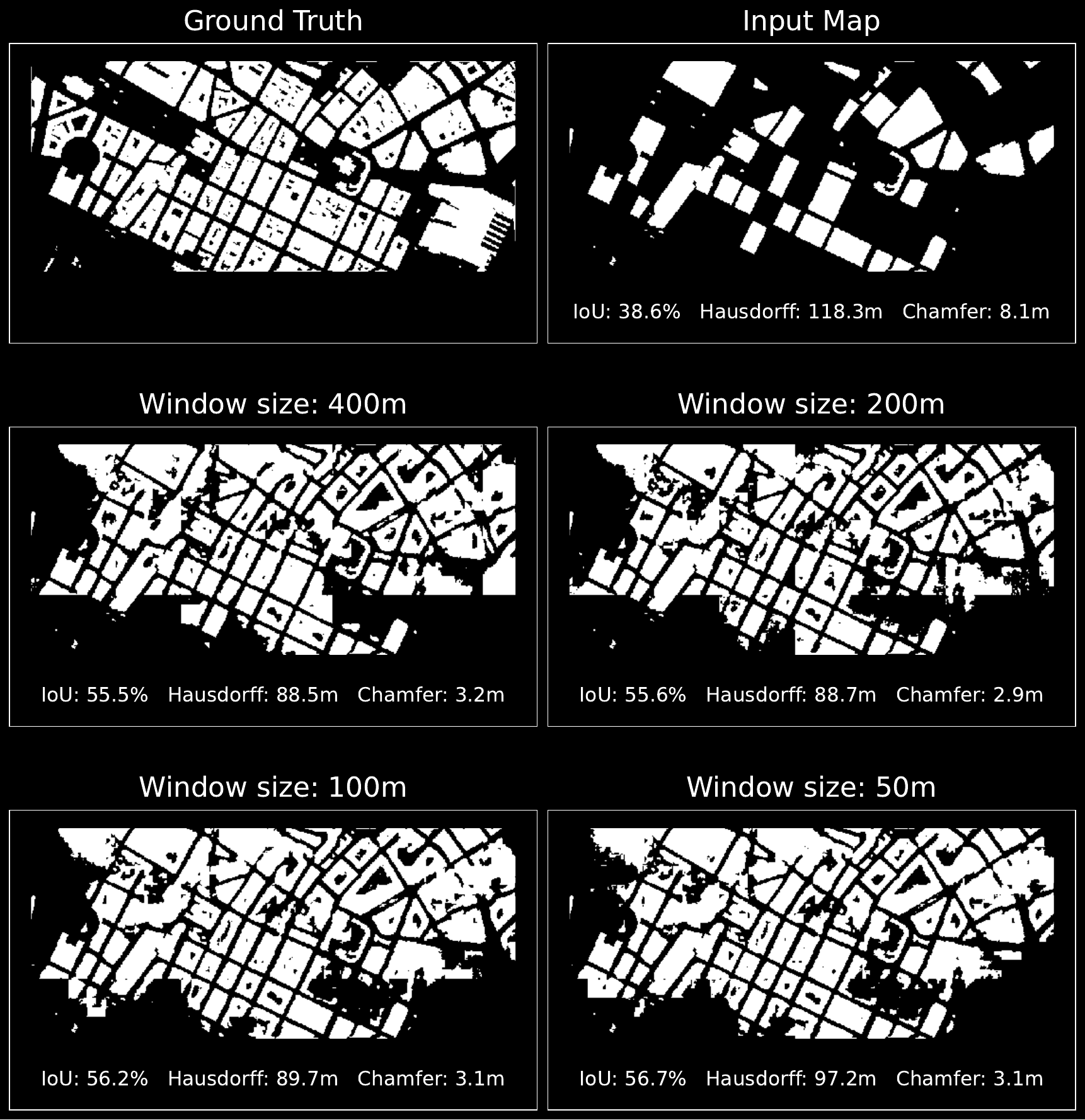}
    \caption{%
        Fused full-extent binary predictions on the Oslo test region under a spatial tiling protocol. The study area is partitioned into \mbox{$400\,\mathrm{m}\times 400\,\mathrm{m}$} tiles and the best Oslo fusion model is applied with overlapping sliding-window inference at window sizes of $400$, $200$, $100$, and $50$ meters; tiles that lack sufficient RF coverage fall back to the corrupted input map. The figure shows the ground-truth map, the corrupted input map, and the four configurations of fusion predictions over the whole Oslo test region. The Macro IoU, Hausdorff, and Chamfer of each configuration are annotated next to the corresponding panel.%
    }
    \label{fig:oslo-tiled-full-region}
\end{figure*}

\subsection{Large-area deployment}
\label{sec:deployment-large}

To complement crop-level evaluation, we assess model's ability to be deployed in a larger area. To that end, we tile the Oslo test area into \mbox{$400\,\mathrm{m}\times 400\,\mathrm{m}$} tiles on a grid anchored at the northwest corner. Overlapping sliding-window inference is performed with window sizes of $400$, $200$, $100$, and $50$ meters. Tiles with insufficient RF support fall back to the corrupted input map. Overlapping tiles are fused by pixel-wise averaging and then binarized at $0.5$. Macro IoU, Hausdorff, and Chamfer distances are computed on the resulting fused map against the Oslo ground truth.

\Cref{fig:oslo-tiled-full-region} summarizes the best Oslo fusion model under this protocol. As a reference, directly using the corrupted input map over the whole Oslo test region yields $38.6\%$ Macro IoU, $118.3\,\mathrm{m}$ Hausdorff, and $8.1\,\mathrm{m}$ Chamfer, which the tiled deployment improves substantially at every window size. At the coarsest inference window of $400\,\mathrm{m}$, the fused prediction reaches $55.5\%$ Macro IoU with Hausdorff $88.5\,\mathrm{m}$ and Chamfer $3.2\,\mathrm{m}$. Macro IoU increases monotonically as the inference window shrinks, reaching $55.6\%$ at $200\,\mathrm{m}$, $56.2\%$ at $100\,\mathrm{m}$, and $56.7\%$ at $50\,\mathrm{m}$, with diminishing returns below $100\,\mathrm{m}$. Chamfer is best at the $200\,\mathrm{m}$ window ($2.9\,\mathrm{m}$) and slightly larger at both coarser and finer windows. Hausdorff is minimized at the $400\,\mathrm{m}$ window ($88.5\,\mathrm{m}$) and mildly worsens at $50\,\mathrm{m}$ ($97.2\,\mathrm{m}$).

For a region of size \mbox{$H\times W$} meters and a window size of \mbox{$s_w$} meters, an overlapping grid would require \mbox{$\bigl\lceil \frac{HW}{s_w^2}\bigr\rceil$} tile evaluations. If each window evaluation costs \mbox{$\Delta(t)$} time, the overall complexity scales as \mbox{$\bigl\lceil \frac{HW}{s_w^2}\bigr\rceil\cdot\Delta(t)$}, with overlap aggregation time being negligible. In our experiemts, we found that the inefrence time $\Delta(t)$ is $329 \pm 88$ milliseconds for a single sample with no batching on an NVIDIA A6000 GPU.

\section{Conclusions}
\label{sec::conclusion}

We explored the challenging task of refining building mapping using RF information and maps provided by open-source mapping platforms. To achieve this, we introduced a novel methodology for corrupting ground-truth maps to simulate real-world OSM inaccuracies. Our approach leveraged two levels of radio information granularity: An angle-based representation and the aggregated path loss. To enhance realism, we incorporated noise into the angle-based representation to better approximate real-world conditions.

To address the problem, we proposed the MapRadioFormer network, a vision transformer-based solution. We conducted a comparative analysis between our deep learning models and a heuristic algorithm, as well as single-modality baselines. Among these baselines, we evaluated an existing neural network designed for map reconstruction using only angle-based RF information. Our experimental results demonstrate that the MapRadioFormer model outperforms the CLIP+UPerNet network \cite{reconstructioniwcmc}, even under the presence of noise introduced to simulate real-world conditions. By integrating both the corrupted map and RF data, our best-performing model achieves a macro IoU of $65.3\%$ on 1,000 test samples. Finally, we conducted a detailed evaluation to analyze the impact of different corruption types on model performance, providing insights into the robustness and effectiveness of our approach.

\section{Future Directions}
\label{sec:future}
A natural extension of our work would be to source inaccurate maps directly from OSM and obtain ground-truth labels from land registry records. Within this framework, two approaches can be considered for acquiring RF data: (i) incorporating the ground-truth maps into a simulation environment to generate synthetic radio measurements or (ii) collecting real-world RF data through field experiments. In particular, improving RF-based performance on Oslo-like real datasets may require both richer real RF information, including metadata such as carrier frequency, and much larger and specifically calibrated synthetic pretraining corpora better matched to real deployments.

An additional research direction involves introducing intrinsic biases regarding building structures to encode prior knowledge about typical building contours, thereby refining the reconstruction process. Another important question, not addressed in this study, is determining an upper bound on performance. Specifically, identifying the maximum achievable IoU given a fixed number of UE-BS pairs. Conversely, it would also be valuable to investigate the minimum amount of RF data required for an accurate reconstruction of the environment. 

Further research could focus on designing neural architectures better suited for the environment mapping task. Another promising direction is the fusion of multiple data modalities to enhance performance. For example, integrating aerial imagery alongside RF and map data could provide complementary spatial information, and evaluating its contribution to the overall mapping accuracy remains an open challenge. Addressing this challenge will require collection of multi-modal datasets of urban environments.

Beyond building footprints, open maps contain many other semantically static object classes, such as roads, railways, and trees. Extending radio-assisted map correction to such layers would require multi-label or multi-task prediction, corruption models tailored to each geometry type, and datasets that co-register those objects with RF observations. We therefore treat generalization to additional static map objects as a concrete direction for future work.

Finally, in real-world settings, except for stationary buildings, the environment often includes dynamic objects such as vehicles, which introduce transient RF scatterers. Extending the concept presented in this work to such environments would require temporal modeling to separate persistent structures from short-lived features. Static buildings could therefore be reliably mapped by aggregating RF–map features across time, while moving vehicles could be detected as transient elements. This extension would not only enhance the robustness of urban mapping but also enable complementary applications in traffic monitoring and vehicular communications.

\section*{Acknowledgements}

This work was supported by funding under the bilateral agreement between CNR (Italy) and HESC MESCS RA (Armenia) as part of the DeepRF project for the 2025–2026 biennium, and by the HESC MESCS RA grant No. 22rl-052 (DISTAL).


\bibliographystyle{elsarticle-harv}
\bibliography{refs}

@article{locadhoc,
  title    = {Deep learning with synthetic data for wireless NLOS positioning with a single base station},
  journal  = {Ad Hoc Networks},
  volume   = {167},
  pages    = {103696},
  year     = {2025},
  issn     = {1570-8705},
  doi      = {https://doi.org/10.1016/j.adhoc.2024.103696},
  url      = {https://www.sciencedirect.com/science/article/pii/S157087052400307X},
  author   = {Hrant Khachatrian and Rafayel Mkrtchyan and Theofanis P. Raptis}
}

@inproceedings{locbds,
  author    = {Darbinyan, Rafayel and Khachatrian, Hrant and Mkrtchyan, Rafayel and Raptis, Theofanis P.},
  booktitle = {2023 IEEE Ninth International Conference on Big Data Computing Service and Applications (BigDataService)},
  title     = {ML-based Approaches for Wireless NLOS Localization: Input Representations and Uncertainty Estimation},
  year      = {2023},
  volume    = {},
  number    = {},
  pages     = {87-94},
  doi       = {10.1109/BigDataService58306.2023.00019}
}

@inproceedings{reconstructioniwcmc,
  author    = {Khachatrian, Hrant and Mkrtchyan, Rafayel and Raptis, Theofanis P.},
  booktitle = {2024 International Wireless Communications and Mobile Computing (IWCMC)},
  title     = {Outdoor Environment Reconstruction with Deep Learning on Radio Propagation Paths},
  year      = {2024},
  volume    = {},
  number    = {},
  pages     = {1498-1503},
  doi       = {10.1109/IWCMC61514.2024.10592367}
}

@article{vit,
  title   = {An image is worth 16x16 words: Transformers for image recognition at scale},
  author  = {Dosovitskiy, Alexey},
  journal = {arXiv preprint arXiv:2010.11929},
  year    = {2020}
}

@article{dinov2,
  title   = {Dinov2: Learning robust visual features without supervision},
  author  = {Oquab, Maxime and Darcet, Timoth{\'e}e and Moutakanni, Th{\'e}o and Vo, Huy and Szafraniec, Marc and Khalidov, Vasil and Fernandez, Pierre and Haziza, Daniel and Massa, Francisco and El-Nouby, Alaaeldin and others},
  journal = {arXiv preprint arXiv:2304.07193},
  year    = {2023}
}

@article{indoorrfmap,
  author     = {Khan, Usman Mahmood and Venkatnarayan, Raghav H. and Shahzad, Muhammd},
  title      = {Using RF Signals to Generate Indoor Maps},
  year       = {2023},
  issue_date = {February 2023},
  publisher  = {Association for Computing Machinery},
  address    = {New York, NY, USA},
  volume     = {19},
  number     = {1},
  issn       = {1550-4859},
  url        = {https://doi.org/10.1145/3534121},
  doi        = {10.1145/3534121},
  journal    = {ACM Trans. Sen. Netw.},
  month      = jan,
  articleno  = {12},
  numpages   = {30}
}

@article{Herfort2023,
  author  = {Herfort, Benjamin and Lautenbach, Sven and Porto de Albuquerque, Jo{\~a}o and Anderson, Jennings and Zipf, Alexander},
  title   = {A spatio-temporal analysis investigating completeness and inequalities of global urban building data in {OpenStreetMap}},
  journal = {Nature Communications},
  year    = {2023},
  volume  = {14},
  number  = {1},
  pages   = {3985},
  month   = {7},
  doi     = {10.1038/541467-023-39698-6},
  issn    = {2041-1723},
  url     = {https://doi.org/10.1038/s41467-023-39698-6}
}

@article{Brovelli2018,
  author         = {Brovelli, Maria Antonia and Zamboni, Giorgio},
  title          = {A New Method for the Assessment of Spatial Accuracy and Completeness of OpenStreetMap Building Footprints},
  journal        = {ISPRS International Journal of Geo-Information},
  volume         = {7},
  year           = {2018},
  number         = {8},
  article-number = {289},
  url            = {https://www.mdpi.com/2220-9964/7/8/289},
  issn           = {2220-9964},
  doi            = {10.3390/ijgi7080289}
}

@article{Hecht2013,
  author  = {Hecht, Robert and Kunze, Carola and Hahmann, Stefan},
  title   = {Measuring Completeness of Building Footprints in OpenStreetMap over Space and Time},
  journal = {ISPRS International Journal of Geo-Information},
  volume  = {2},
  year    = {2013},
  number  = {4},
  pages   = {1066--1091},
  url     = {https://www.mdpi.com/2220-9964/2/4/1066},
  issn    = {2220-9964},
  doi     = {10.3390/ijgi2041066}
}

@article{ruble2018wireless,
  title     = {Wireless localization for mmWave networks in urban environments},
  author    = {Ruble, Macey and G{\"u}ven{\c{c}}, Ismail},
  journal   = {EURASIP journal on advances in signal processing},
  volume    = {2018},
  pages     = {1--19},
  year      = {2018},
  publisher = {Springer}
}

@inproceedings{indoorsar,
  author    = {Sévigny, Pascale and DiFilippo, David J.},
  booktitle = {2013 IEEE Radar Conference (RadarCon13)},
  title     = {A multi-look fusion approach to through-wall radar imaging},
  year      = {2013},
  volume    = {},
  number    = {},
  pages     = {1-6},
  doi       = {10.1109/RADAR.2013.6586154}
}

@article{SIRE,
  author   = {Le, Calvin and Dogaru, Traian and Nguyen, Lam and Ressler, Marc A.},
  journal  = {IEEE Transactions on Geoscience and Remote Sensing},
  title    = {Ultrawideband (UWB) Radar Imaging of Building Interior: Measurements and Predictions},
  year     = {2009},
  volume   = {47},
  number   = {5},
  pages    = {1409-1420},
  doi      = {10.1109/TGRS.2009.2016653}
}

@article{mobileOFDM,
  author   = {Baquero Barneto, Carlos and Rastorgueva-Foi, Elizaveta and Keskin, Musa Furkan and Riihonen, Taneli and Turunen, Matias and Talvitie, Jukka and Wymeersch, Henk and Valkama, Mikko},
  journal  = {IEEE Transactions on Vehicular Technology},
  title    = {Millimeter-Wave Mobile Sensing and Environment Mapping: Models, Algorithms and Validation},
  year     = {2022},
  volume   = {71},
  number   = {4},
  pages    = {3900-3916},
  doi      = {10.1109/TVT.2022.3146003}
}

@inproceedings{emreflectors,
  author    = {Xie, Chenxi and Ma, Shanshan and Zhou, Bingpeng},
  booktitle = {2023 IEEE 11th International Conference on Information, Communication and Networks (ICICN)},
  title     = {Environment Mapping Based on mmWave MIMO OFDM Communication Systems towards 6G Integrated Communication and Sensing},
  year      = {2023},
  volume    = {},
  number    = {},
  pages     = {194-200},
  doi       = {10.1109/ICICN59530.2023.10393672}
}

@inproceedings{3dwifirssiuav,
  title     = {3D through-wall imaging with unmanned aerial vehicles using WiFi},
  author    = {Karanam, Chitra R and Mostofi, Yasamin},
  booktitle = {Proceedings of the 16th ACM/IEEE International Conference on Information Processing in Sensor Networks},
  pages     = {131--142},
  year      = {2017}
}

@misc{OpenStreetMap,
  author       = {{OpenStreetMap contributors}},
  title        = {{Planet dump retrieved from https://planet.osm.org }},
  howpublished = {\url{ https://www.openstreetmap.org }},
  year         = {2017}
}

@article{wair-d,
  title   = {Wair-d: Wireless ai research dataset},
  author  = {Huangfu, Yourui and Wang, Jian and Dai, Shengchen and Li, Rong and Wang, Jun and Huang, Chongwen and Zhang, Zhaoyang},
  journal = {arXiv preprint arXiv:2212.02159},
  year    = {2022}
}

@techreport{3gpp,
  author      = {3GPP},
  title       = {3rd Generation Partnership Project; Technical Specification Group Radio Access Network; Requirements for Evolved UTRA (E-UTRA) and Evolved UTRAN (E-UTRAN) (Release 7)},
  institution = {3GPP},
  year        = {2006},
  number      = {TR 25.913 V7.3.0 (2006-03)}
}

@article{sionna,
  title   = {Sionna: An Open-Source Library for Next-Generation Physical Layer Research},
  author  = {Hoydis, Jakob and Cammerer, Sebastian and {Ait Aoudia}, Fayçal and Vem, Avinash and Binder, Nikolaus and Marcus, Guillermo and Keller, Alexander},
  year    = {2022},
  month   = {Mar.},
  journal = {arXiv preprint},
  online  = {https://arxiv.org/abs/2203.11854}
}

@article{indoorpathloss2,
  title   = {IPP-Net: A Generalizable Deep Neural Network Model for Indoor Pathloss Radio Map Prediction},
  author  = {Feng, Bin and Zheng, Meng and Liang, Wei and Zhang, Lei},
  journal = {arXiv preprint arXiv:2501.06414},
  year    = {2025}
}

@article{indoorpathloss4,
  title   = {TransPathNet: A Novel Two-Stage Framework for Indoor Radio Map Prediction},
  author  = {Li, Xin and Liu, Ran and Xu, Saihua and Razul, Sirajudeen Gulam and Yuen, Chau},
  journal = {arXiv preprint arXiv:2501.16023},
  year    = {2025}
}

@article{indoorpathloss7,
  title  = {Indoor Path Loss Prediction Leveraging Radio Tomographic Maps},
  author = {Viet, Pham Q and Romero, Daniel},
  year   = {2024}
}

@article{indoorpathloss8,
  title   = {Vision Transformers for Efficient Indoor Pathloss Radio Map Prediction},
  author  = {Ghukasyan, Edvard and Khachatrian, Hrant and Mkrtchyan, Rafayel and Raptis, Theofanis P},
  journal = {arXiv preprint arXiv:2412.09507},
  year    = {2024}
}

@inproceedings{outdoorpathloss2,
  title        = {Agile radio map prediction using deep learning},
  author       = {Krijestorac, Enes and Sallouha, Hazem and Sarkar, Shamik and Cabric, Danijela},
  booktitle    = {ICASSP 2023-2023 IEEE International Conference on Acoustics, Speech and Signal Processing (ICASSP)},
  pages        = {1--2},
  year         = {2023},
  organization = {IEEE}
}

@inproceedings{outdoorpathloss3,
  title        = {Deep learning-based path loss prediction for outdoor wireless communication systems},
  author       = {Qiu, Kehai and Bakirtzis, Stefanos and Song, Hui and Wassell, Ian and Zhang, Jie},
  booktitle    = {ICASSP 2023-2023 IEEE International Conference on Acoustics, Speech and Signal Processing (ICASSP)},
  pages        = {1--2},
  year         = {2023},
  organization = {IEEE}
}

@article{pathlosstransformer,
  title   = {Radionet: Transformer based radio map prediction model for dense urban environments},
  author  = {Tian, Yu and Yuan, Shuai and Chen, Weisheng and Liu, Naijin},
  journal = {arXiv preprint arXiv:2105.07158},
  year    = {2021}
}

@article{pathlossunet,
  title     = {RadioUNet: Fast radio map estimation with convolutional neural networks},
  author    = {Levie, Ron and Yapar, {\c{C}}a{\u{g}}kan and Kutyniok, Gitta and Caire, Giuseppe},
  journal   = {IEEE Transactions on Wireless Communications},
  volume    = {20},
  number    = {6},
  pages     = {4001--4015},
  year      = {2021},
  publisher = {IEEE}
}

@article{loc1,
  title     = {Deep neural networks for wireless localization in indoor and outdoor environments},
  author    = {Zhang, Wei and Liu, Kan and Zhang, Weidong and Zhang, Youmei and Gu, Jason},
  journal   = {Neurocomputing},
  volume    = {194},
  pages     = {279--287},
  year      = {2016},
  publisher = {Elsevier}
}

@article{loc2,
  title     = {MetaLoc: Learning to learn wireless localization},
  author    = {Gao, Jun and Wu, Dongze and Yin, Feng and Kong, Qinglei and Xu, Lexi and Cui, Shuguang},
  journal   = {IEEE Journal on Selected Areas in Communications},
  year      = {2023},
  publisher = {IEEE}
}

@inproceedings{loc3,
  title        = {WiDeep: WiFi-based accurate and robust indoor localization system using deep learning},
  author       = {Abbas, Moustafa and Elhamshary, Moustafa and Rizk, Hamada and Torki, Marwan and Youssef, Moustafa},
  booktitle    = {2019 IEEE International Conference on Pervasive Computing and Communications (PerCom},
  pages        = {1--10},
  year         = {2019},
  organization = {IEEE}
}

@article{loc4,
  title     = {UWB indoor localization using deep learning LSTM networks},
  author    = {Poulose, Alwin and Han, Dong Seog},
  journal   = {Applied Sciences},
  volume    = {10},
  number    = {18},
  pages     = {6290},
  year      = {2020},
  publisher = {MDPI}
}

@article{aerial1,
  title     = {A deep learning approach to an enhanced building footprint and road detection in high-resolution satellite imagery},
  author    = {Ayala, Christian and Sesma, Rub{\'e}n and Aranda, Carlos and Galar, Mikel},
  journal   = {Remote Sensing},
  volume    = {13},
  number    = {16},
  pages     = {3135},
  year      = {2021},
  publisher = {MDPI}
}

@article{vargas2019correcting,
  title     = {Correcting rural building annotations in OpenStreetMap using convolutional neural networks},
  author    = {Vargas-Mu{\~n}oz, John E and Lobry, Sylvain and Falc{\~a}o, Alexandre X and Tuia, Devis},
  journal   = {ISPRS journal of photogrammetry and remote sensing},
  volume    = {147},
  pages     = {283--293},
  year      = {2019},
  publisher = {Elsevier}
}

@article{li2022improving,
  title     = {Improving OpenStreetMap missing building detection using few-shot transfer learning in sub-Saharan Africa},
  author    = {Li, Hao and Herfort, Benjamin and Lautenbach, Sven and Chen, Jiaoyan and Zipf, Alexander},
  journal   = {Transactions in GIS},
  volume    = {26},
  number    = {8},
  pages     = {3125--3146},
  year      = {2022},
  publisher = {Wiley Online Library}
}

@inproceedings{li2023rethink,
  title     = {Rethink geographical generalizability with unsupervised self-attention model ensemble: A case study of openstreetmap missing building detection in africa},
  author    = {Li, Hao and Wang, Jiapan and Zollner, Johann Maximilian and Mai, Gengchen and Lao, Ni and Werner, Martin},
  booktitle = {Proceedings of the 31st ACM International Conference on Advances in Geographic Information Systems},
  pages     = {1--9},
  year      = {2023}
}

@article{chen2023land,
  title   = {Land-cover change detection using paired openstreetmap data and optical high-resolution imagery via object-guided transformer},
  author  = {Chen, Hongruixuan and Lan, Cuiling and Song, Jian and Broni-Bediako, Clifford and Xia, Junshi and Yokoya, Naoto},
  journal = {arXiv preprint arXiv:2310.02674},
  year    = {2023}
}

@inproceedings{laddha2016map,
  title        = {Map-supervised road detection},
  author       = {Laddha, Ankit and Kocamaz, Mehmet Kemal and Navarro-Serment, Luis E and Hebert, Martial},
  booktitle    = {2016 IEEE Intelligent Vehicles Symposium (IV)},
  pages        = {118--123},
  year         = {2016},
  organization = {IEEE}
}

@article{lecun1998gradient,
  title     = {Gradient-based learning applied to document recognition},
  author    = {LeCun, Yann and Bottou, L{\'e}on and Bengio, Yoshua and Haffner, Patrick},
  journal   = {Proceedings of the IEEE},
  volume    = {86},
  number    = {11},
  pages     = {2278--2324},
  year      = {1998},
  publisher = {Ieee}
}

@book{hausdorff_grundzuge_1914,
  series    = {Göschens {Lehrbücherei}/{Gruppe} {I}: {Reine} und {Angewandte} {Mathematik} {Series}},
  title     = {Grundzüge der {Mengenlehre}},
  isbn      = {978-3-11-098985-4},
  publisher = {Von Veit},
  author    = {Hausdorff, F.},
  year      = {1914}
}

@inproceedings{10.5555/1622943.1622971,
  author    = {Barrow, H. G. and Tenenbaum, J. M. and Bolles, R. C. and Wolf, H. C.},
  title     = {Parametric correspondence and chamfer matching: two new techniques for image matching},
  year      = {1977},
  publisher = {Morgan Kaufmann Publishers Inc.},
  address   = {San Francisco, CA, USA},
  booktitle = {Proceedings of the 5th International Joint Conference on Artificial Intelligence - Volume 2},
  pages     = {659–663},
  numpages  = {5},
  location  = {Cambridge, USA},
  series    = {IJCAI'77}
}

@article{SOHN200743,
  title    = {Data fusion of high-resolution satellite imagery and LiDAR data for automatic building extraction},
  journal  = {ISPRS Journal of Photogrammetry and Remote Sensing},
  volume   = {62},
  number   = {1},
  pages    = {43-63},
  year     = {2007},
  issn     = {0924-2716},
  doi      = {https://doi.org/10.1016/j.isprsjprs.2007.01.001},
  url      = {https://www.sciencedirect.com/science/article/pii/S0924271607000032},
  author   = {Gunho Sohn and Ian Dowman}
}

@article{CESARIO2022101687,
  title    = {Multi-density urban hotspots detection in smart cities: A data-driven approach and experiments},
  journal  = {Pervasive and Mobile Computing},
  volume   = {86},
  pages    = {101687},
  year     = {2022},
  issn     = {1574-1192},
  doi      = {https://doi.org/10.1016/j.pmcj.2022.101687},
  url      = {https://www.sciencedirect.com/science/article/pii/S1574119222001018},
  author   = {Eugenio Cesario and Paschal I. Uchubilo and Andrea Vinci and Xiaotian Zhu}
}

@article{BERALDI2020101221,
  title    = {Distributed load balancing for heterogeneous fog computing infrastructures in smart cities},
  journal  = {Pervasive and Mobile Computing},
  volume   = {67},
  pages    = {101221},
  year     = {2020},
  issn     = {1574-1192},
  doi      = {https://doi.org/10.1016/j.pmcj.2020.101221},
  url      = {https://www.sciencedirect.com/science/article/pii/S1574119220300791},
  author   = {Roberto Beraldi and Claudia Canali and Riccardo Lancellotti and Gabriele Proietti Mattia}
}

@article{zheng2022radiocycle,
  title     = {{Radiocycle: Deep dual learning based radio map estimation}},
  author    = {Zheng, Yi and Zhang, Tianqian and Liao, Cunyi and Wang, Ji and Liu, Shouyin},
  journal   = {{KSII Transactions on Internet and Information Systems (TIIS)}},
  volume    = {{16}},
  number    = {11},
  pages     = {{3780--3797}},
  year      = {2022},
  publisher = {Korean Society for Internet Information}
}

@article{zhang2021building,
  title     = {{Building layout tomographic reconstruction via commercial WiFi signals}},
  author    = {Zhang, Yang and Chen, Jiahui and Guo, Shisheng and Yang, Xiaobo and Cui, Guolong},
  journal   = {{IEEE Internet of Things Journal}},
  volume    = {{8}},
  number    = {20},
  pages     = {{15500--15511}},
  year      = {2021},
  publisher = {IEEE}
}

@inproceedings{peng20223d,
  title        = {{3D city map reconstruction from LEO communication satellite SNR measurements}},
  author       = {Peng, Bohan and Jing, Wenpeng and Zheng, Ziyuan and Wen, Xiangming and Lu, Zhaoming and Wang, Zhifei},
  booktitle    = {{2022 IEEE/CIC International Conference on Communications in China (ICCC)}},
  pages        = {{268--273}},
  year         = {2022},
  organization = {IEEE}
}

@article{wang2025bit,
  title     = {{Bit to brick: from cellular mobile signals to 3D city map creation}},
  author    = {Wang, Yu and Basiri, Anahid},
  journal   = {{Big Earth Data}},
  pages     = {{1--25}},
  year      = {2025},
  publisher = {Taylor \& Francis}
}

@article{mukherjee2019losi,
  title     = {{LoSI: Large scale location inference through FM signal integration and estimation}},
  author    = {Mukherjee, Tathagata and Kumar, Piyush and Pati, Debdeep and Blasch, Erik and Pasiliao, Eduardo and Xu, Liqin},
  journal   = {{Big Data Mining and Analytics}},
  volume    = {{2}},
  number    = {4},
  pages     = {{319--348}},
  year      = {2019},
  publisher = {TUP}
}

@Article{s23094266,
AUTHOR = {De Nardis, Luca and Caso, Giuseppe and Alay, Ozgu and Neri, Marco and Brunstrom, Anna and Di Benedetto, Maria-Gabriella},
TITLE = {Positioning by Multicell Fingerprinting in Urban NB-IoT Networks},
JOURNAL = {Sensors},
VOLUME = {23},
YEAR = {2023},
NUMBER = {9},
ARTICLE-NUMBER = {4266},
URL = {https://www.mdpi.com/1424-8220/23/9/4266},
PubMedID = {37177470},
ISSN = {1424-8220},
DOI = {10.3390/s23094266}
}

\balance






\end{document}